%% file: 00_main.tex
\documentclass[manuscript, screen]{acmart}

\usepackage{amsmath,amsfonts}
\usepackage{array}
\usepackage[caption=false,font=normalsize,labelfont=sf,textfont=sf]{subfig}
\usepackage{textcomp}
\usepackage{stfloats}
\usepackage{verbatim}
\usepackage[dvipsnames]{xcolor}
\usepackage{graphicx}
\usepackage{hyperref}
\usepackage{xspace}
\usepackage{enumitem}
\usepackage{tabularx}
\usepackage[normalem]{ulem}
\usepackage[table]{colortbl}
\usepackage{textcomp}
\usepackage{standalone}
\usepackage{booktabs}
\usepackage{pgfplots}
\pgfplotsset{compat=newest}
\usepackage{algorithm} 
\usepackage{algpseudocode} 
\definecolor{LightCyan}{rgb}{0.88,1,1}

\newcommand{\sysname}{PEARL\xspace}

\AtBeginDocument{%
  }

\setcopyright{none}

\copyrightyear{2026}
\acmYear{2026}
\acmDOI{XXXXXXX.XXXXXXX}
\acmConference[ICCPS '26]{ACM/IEEE International Conference on Cyber-Physical Systems (ICCPS) }{May 11--24,
  2026}{Saint Malo, France}

\renewcommand\footnotetextcopyrightpermission[1]{} 

\begin{document}


\title{\sysname:  Structural Privacy-Utility Control in Human-Centric CPS via Personalized Early-Exit Deep Reinforcement Learning }

\author{Mojtaba Taherisadr}
\email{taherisa@uci.edu}
\author{Salma Elmalaki}
\email{salma.elmalaki@uci.edu}
\affiliation{%
   \institution{University of California, Irvine}
   \country{USA}
}








\renewcommand{\shortauthors}{Taherisadr et al.}

\begin{abstract}

In the evolving landscape of human-centric Cyber-Physical Systems (CPS), personalized privacy solutions are becoming increasingly crucial due to the dynamic and diverse nature of human interactions. Traditional static privacy models often fail to meet the changing privacy needs of individual users. In human-centric CPS, personalized Deep Reinforcement Learning (DRL) agents must share fine-grained control actions with cloud services, inadvertently exposing sensitive private states to inference attacks by honest-but-curious adversaries. This paper introduces \sysname (Personalized Early-exit Adaptive Reinforcement Learning), a novel framework that addresses this challenge through structural privacy control rather than data perturbation. \sysname's core innovation is a dual-path Early-Exit Deep Q-Network (EE-DQN) deployed at the edge, which uses Mutual Information (MI) between private states and observable actions as a principled leakage metric to train per-branch binary labels: Utility Confidence Labels (UCL), which verify that a branch's action quality remains within an acceptable fraction of optimal, and Privacy Confidence Labels (PCL), which verify that the branch's MI-based leakage remains below a user-defined risk threshold. At inference time, \sysname dynamically selects the shallowest exit branch that simultaneously satisfies both UCL and PCL, structurally limiting the descriptive power of shared actions without injecting noise into the control loop. An MI-based monitoring feedback loop continuously tracks behavioral drift and triggers adaptive retraining when the deployed policy's privacy-utility profile shifts significantly, ensuring robustness to human variability over time. We validate \sysname across two distinct human-centric CPS domains: a personalized smart-home HVAC system and a VR smart classroom. Empirical results demonstrate that \sysname reduces adversarial state-inference accuracy by an average of 25.67\% across both applications, while incurring a controlled utility cost of 10–16\%, confirming a practical, personalized, and dynamically enforceable privacy-utility tradeoff.
\end{abstract}



\keywords{Human-centric CPS, Reinforcement Learning, Early-Exit, Privacy, Personalization}


\maketitle

\input{01_introduction}

\input{02_related}
\input{03_threatmodel}

\input{04_privacymetric}

\input{05_algorithm}

\input{06_application1_home}

\input{07_application2_learning}

\input{08_discussion}

\input{09_conclusion}


\section*{Acknowledgment}
This research was partially supported by the National Science Foundation (NSF) award $2339266$.

\bibliographystyle{ACM-Reference-Format}
\bibliography{references}

\end{document}

%% file: 01_introduction.tex
\vspace{-2mm}
\section{Introduction}
The increasing ubiquity of intelligent devices has fundamentally reshaped the architecture of traditional Cyber-Physical Systems (CPS), driving the domain toward solving grand societal-scale challenges and focusing on the core concept of Cyber-Physical-Human Systems (CPHS). This paradigm recognizes the blurring boundaries between human users and technology in highly complex systems, ranging from robotic assistance devices to smart homes and autonomous vehicles~\cite{IEEE_CSS_Roadmap_2023}. Consequently, the next generation of control systems must prioritize Human-Centered CPS design to effectively manage this interaction. A critical challenge emerging in this human-centric landscape is maintaining safety and establishing trust  while simultaneously managing the privacy-utility tradeoff for individual users. The high degree of uncertainty and variability inherent in human behavior renders conventional, static privacy models obsolete, creating a pressing need for principled approaches that can strike an effective balance between system robustness and adaptivity to individual dynamics and preferences~\cite{taherisadr2023adaparl}. It is only through such personalized, dynamic solutions that systems can respect human privacy and ensure the widespread, safe deployment of these technologies.

To address these fundamental challenges in the CPHS domain, this paper focuses on mitigating privacy leaks in machine learning-based human-centric CPS. We specifically examine the use of Deep Reinforcement Learning (DRL) within this context. RL is a natural fit for this context due to its inherent capability to adapt to dynamic human intentions and responses (inter- and intra-human variability) in real-time environments, which static control policies struggle to manage effectively \cite{sadigh2017active, hadfield2016cooperative, elmalaki2018sentio, ahadi2021adas}.

While DRL's power to personalize CPS for optimal individual experiences is well-documented, its reliance on continuous, fine-grained observation of user behavior creates a significant drawback: the potential to access and infer highly sensitive private human states~\cite{omonkhoa2021review}. This issue is particularly critical in safety-critical CPS dealing with delicate personal information—such as those found in healthcare, finance, and autonomous vehicles—where the persistent collection and processing of user data pose acute privacy concerns \cite{hlavka2020security}. Therefore, the core challenge is not merely achieving personalization, but dynamically balancing the personalized utility gained against the individual privacy risk incurred within the decision-making control loop of a CPS.

Previous research has explored privacy vulnerabilities in RL and control systems, often focusing on establishing the privacy-utility tradeoff using techniques like differential privacy or adversarial training \cite{taherisadr2023adaparl, wang2019privacy}. However, these approaches often yield static or sub-optimally generalizable policies. They fail to account for the minute-to-minute changes and the inherent individual differences (variability) that define the human component in modern CPHS. Recognizing this gap, we propose PEARL (Personalized Early-exit Adaptive Reinforcement Learning), a novel DRL approach that leverages an early-exit strategy to directly address the personalized and dynamic nature of the problem. This early-exit mechanism allows the CPS control system to make a privacy-preserving decision as soon as enough data is observed, thereby minimizing the exposure window and tailoring the privacy-utility tradeoff to the unique variability of each human user in real-time.

Our contributions can be summarized as follows:
\begin{enumerate}[noitemsep, leftmargin=*,topsep=0pt] 
\item \textbf{A Novel Dynamic Privacy-Utility Framework:} We introduce \sysname, a framework to leverage an Early-Exit Deep Q-Network (EE-DQN) to dynamically balance the personalized privacy-utility tradeoff. The agent's decision is constrained by user-defined privacy and utility preferences.

\item \textbf{Adaptive Monitoring and Retraining Loop:} We propose a mechanism using for real-time risk assessment. This novel feedback loop automatically triggers the re-training of the EE-DQN agent to incorporate human variability and ensure policy robustness.

\item \textbf{Validation in Human-Centric CPS:} We empirically validate \sysname's dynamic adaptation capabilities in two distinct contexts: Smart Home Environments and VR Smart Classrooms. Our results show that \sysname enhances privacy protection by 31\% on average.



\end{enumerate}

%% file: 02_related.tex
\vspace{-2mm}
\section{Background and Related Work}\label{sec:related}

\paragraph{\textbf{Deep Reinforcement Learning for Human-Centric CPS}}
The increasing complexity of Cyber-Physical-Human Systems (CPHS) demands control policies that can operate robustly while dynamically accommodating the high variability and uncertainty inherent in human behavior. Traditional control policies and static machine learning models often fail to capture the nuances of both inter-human (between different users) and intra-human (evolution over time for one user) variability~\cite{elmalaki2018sentio, ahadi2021adas}. Deep Reinforcement Learning (DRL) offers a compelling solution, leveraging the representational power of deep neural networks to learn optimal control policies in complex, sequential decision-making environments. Specifically, DRL is critical for synthesizing personalized policies that adapt to human intentions and responses in real-time interactions~\cite{sadigh2017active, hadfield2016cooperative}. The need for this fine-grained, personalized control requires continuous observation of sensitive human state data, which, while maximizing system utility, inevitably introduces significant privacy risks~\cite{taherisadr2023adaparl}. This inherent conflict forms the basis of the necessary privacy-utility tradeoff that must be managed in CPHS. Our emphasis on using DQN-based DRL is driven by two critical attributes:
\begin{itemize}[noitemsep, leftmargin=*,topsep=0pt]
\item Computational Complexity and Scalability: RL offers computational scalability advantages compared to other techniques, especially game-theoretic approaches. Many game-theoretic methods for sequential decision-making are recognized as computationally infeasible \cite{fraenkel2004complexity}.

\item Generalizability: RL possesses a unique capacity to directly model the consequences of decisions, leverage temporal feedback during the learning process, and enhance decision-making policy performance across a broad spectrum of systems. This attribute is particularly valuable for human-centric CPS.
\end{itemize}

\vspace{-2mm}

\paragraph{\textbf{Limitations of Current Privacy Techniques in CPHS}} Existing Privacy techniques generally fall into two categories: cryptographic and perturbation-based. 
 \begin{itemize}[noitemsep, leftmargin=*,topsep=0pt] 
 \item \textbf{Cryptographic PETs:} Techniques like Homomorphic Encryption (HE) and Multi-Party Computation (MPC) provide strong confidentiality guarantees for edge-cloud synchronization~\cite{zhang2022homomorphic}. However, their computational latency is often incompatible with the real-time constraints of HCS, where control loops must operate at millisecond intervals.

 \item \textbf{Differential Privacy (DP):} Perturbation-based methods, most notably Differential Privacy (DP), offer a rigorous mathematical framework for bounding information leakage by injecting stochastic noise into gradients or model outputs~\cite{patel2022model, wang2019privacy}. While foundational, static DP formulations typically impose a global, ``one-size-fits-all'' privacy budget that is context-blind. Such approaches fail to accommodate the non-stationary nature of human behavior or the fluctuating risk levels inherent in physical environments—for instance, the varying privacy expectations of a user transitioning from a public smart classroom to a private residence~\cite{taherisadr2023adaparl}. Moreover, recent work in the literature has shown that applying DP in general without understanding human behavior can lead to disparity in privacy risk across individuals~\cite{Zhao2026FinP}.

\end{itemize}

Recent advancements in personalized and adaptive DP~\cite{niu2021adapdp, fallah2020personalized} attempt to address this by tuning noise scales to individual user profiles. However, in the context of CPHS, these methods face a fundamental limitation: any noise added to control signals or state observations directly propagates through the physical plant, potentially compromising control stability and system safety. \sysname shifts this paradigm by utilizing the architecture itself as the privacy mechanism. Rather than obfuscating data through post-hoc perturbation, \sysname leverages structural anonymization; by dynamically selecting an early exit, the system physically truncates the generation of high-fidelity, descriptive feature representations before they can be externalized to the cloud. This approach provides a structural constraint on information leakage, empirically observed to reduce Mutual Information (MI) as exit depth decreases, consistent with theoretical intuition from information processing inequalities~\cite{li2022auditing}. 

\vspace{-2mm}
\paragraph{\textbf{Early-Exit Architectures for Dynamic Policy Control}}
To achieve a dynamic and real-time adjustment of the privacy-utility tradeoff, the control policy must have an architectural means to modulate its information processing depth. Early-Exit (EE) architectures provide this foundation\cite{teerapittayanon2016branchynet}. Originally developed to improve computational efficiency and reduce latency in Deep Neural Networks (DNNs) by terminating inference once a predetermined confidence threshold is met, the EE principle has been applied to DRL for similar purposes~\cite{matsubara2022split}. For instance, multi-exit DRL models have been shown to accelerate inference speed by facilitating rapid policy execution\cite{kosta2022rapid}.

The architectural design of multi-exit networks is highly relevant to privacy because the number of layers processed directly correlates with the amount of information extracted and, consequently, the potential privacy risk incurred~\cite{li2022auditing, wu2023enhancing}. By treating the exit confidence level as a control variable, the network can be trained to dynamically determine the optimal depth of processing. This structural adaptability provides the mechanism to satisfy the dual requirements identified in this work: minimizing loss in performance (utility) while restricting sensitive information processing (privacy). Our proposed system, \sysname, uniquely integrates this Early-Exit architecture into a DRL framework to explicitly manage the confidence-based, personalized dynamic tradeoff between system utility and individual human privacy.

%% file: 03_threatmodel.tex
\vspace{-2mm}\section{Threat Model in Human-Centric Systems}\label{sec:threatmodel}

The increasing computational power and memory capacity of edge devices have enabled the execution of resource-intensive tasks, including reinforcement learning (RL) algorithms, directly at the edge. This capability is crucial for real-time decision-making in ubiquitous environments. However, synchronizing multiple applications often requires centralized decision-making, typically facilitated by a cloud-based server. Examples include the NEST program in Southern California and Southern California Edison's (SCE) Rush Hour Rewards, where cloud servers optimize energy consumption by controlling HVAC setpoints for registered homes~\cite{nestcali,rushh}.

Our privacy framework addresses a threat model situated within a realistic Cyber-Physical-Human System (CPHS) that utilizes a Human-in-the-Loop (HITL), edge-cloud architecture, common in personalized, real-time applications. This architecture is necessitated by the need for real-time local control combined with centralized management. This edge-cloud architecture is common in many pervasive applications~\cite{bovornkeeratiroj2020repel,papst2022share,stirapongsasuti2019decision}.

\begin{figure}[!t]
\centering
\includegraphics[width=0.6\columnwidth]{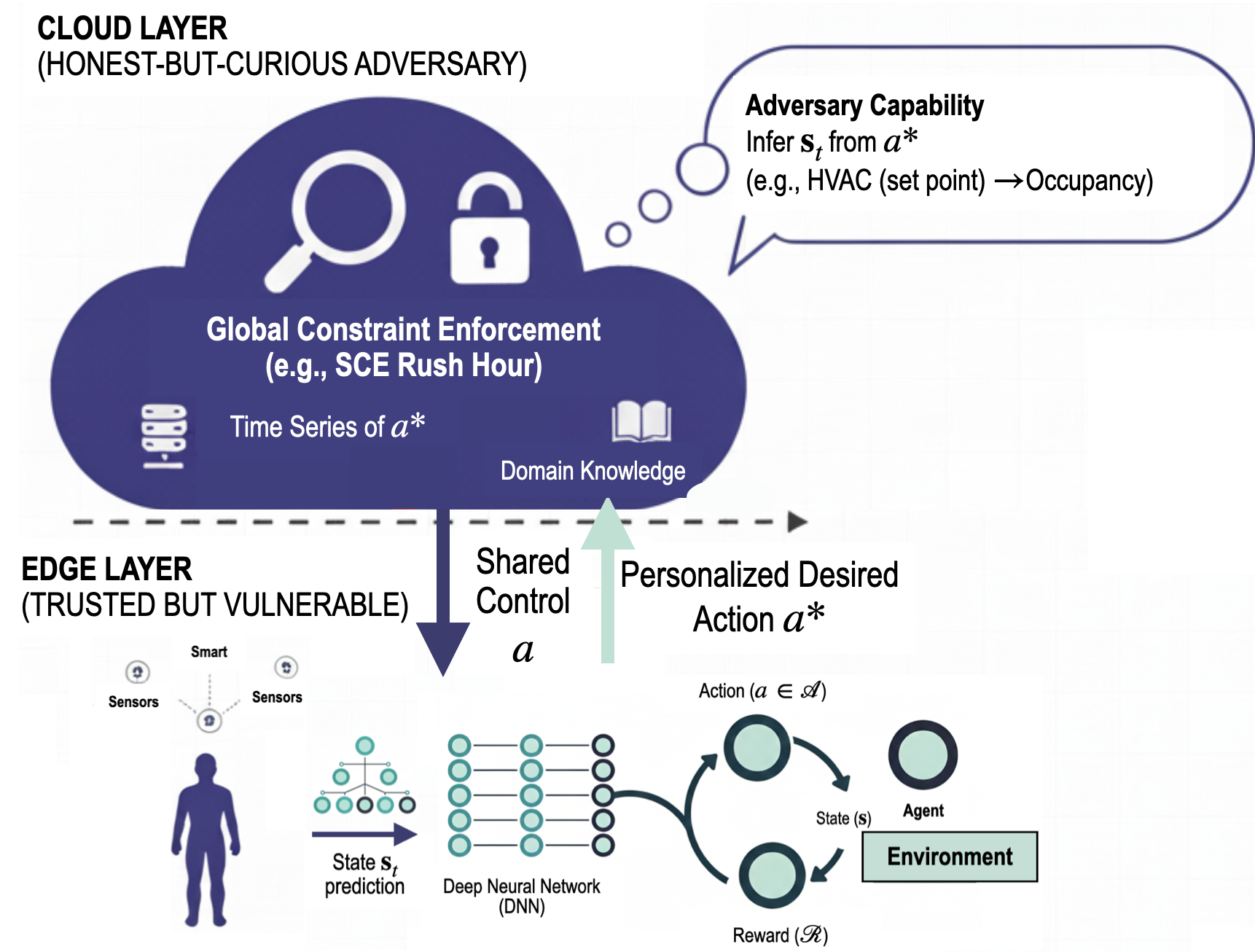}
\caption{Threat Model: A trusted edge device running DQN shares recommended actions with an ``honest-but-curious'' cloud. The cloud, leveraging domain knowledge, may infer sensitive information from the action sequence, compromising user privacy.}
\label{fig:threat}
\end{figure}

The system is defined by two operational levels as shown in Figure~\ref{fig:threat}:
\begin{itemize}[noitemsep, leftmargin=*,topsep=0pt]
\item Edge Layer (Trusted but Vulnerable): Edge devices (e.g., smart sensors) collect sensitive human state data ($s_t$). This layer hosts the local Deep Reinforcement Learning (DRL) agent, which analyzes the personalized data to determine an optimal, adaptive ``Desired Action'' ($a^*$).

\item Cloud Layer (Honest-but-Curious Adversary): The local agent's ($a^*$) must be forwarded to a centralized cloud server to perform essential system functions, such as global constraint enforcement (e.g., SCE program) and application synchronization. This cloud is modeled as an ``honest-but-curious'' cloud service. This model is standard in security literature (e.g., \cite{zhang2022homomorphic, xiao2023micpro}); the adversary honestly executes the action but leverages its access to the shared control channel to attempt sensitive information inference.
\end{itemize}

\vspace{-2mm}
\noindent\paragraph{\textbf{Threat of State Inference from Personalized Action Sequence}}
The core vulnerability arises directly from the DRL agent's need for fine-grained personalization to maximize utility, which makes its action ($a^*$) highly descriptive of the user's private state ($s_t$).
\begin{itemize}[noitemsep, leftmargin=*,topsep=0pt]
\item Attack Target: The data exposed to the adversary is the time-series sequence of the personalized actions ($a^*$) outputted by the edge-based DQN agent.

\item Adversary's Capability: The cloud is assumed to possess unlimited computational resources and application domain knowledge. Given the personalized nature of ($a^*$), the adversary can exploit the strong correlation between the action and the underlying human state. For example, personalized HVAC setpoints ($a^*$) inherently reveal sensitive information about a user's occupancy patterns, thermal preferences, or even activity levels~\cite{taherisadr2023adaparl, elmalaki2019spycon}.
\end{itemize}

This threat is validated by real-world security research demonstrating successful attacks based on sensor and output data. Prior work has shown that systems designed for utility can be repurposed for privacy attacks. For example, side-channel attacks have successfully inferred sensitive information like vital signs and private embedded information from AR/VR motion sensor outputs~\cite{zhang2023facereader} and speech content from microphone outputs intended only for ambient noise analysis~\cite{xiao2023micpro}.

These precedents confirm that any highly successful personalized control signal ($a^*$) is a potent vector for private state leakage, as it must encode unique behavioral patterns to achieve utility. This necessitates a solution that can dynamically decouple utility from privacy risk.

\vspace{-2mm}
\noindent \paragraph{\textbf{Motivation for the Dynamic Early-Exit Defense}} The inherent conflict between the need for personalized utility and the risk of action-based state inference necessitates a dynamic defense. Conventional privacy mechanisms are limited:
\begin{itemize}[noitemsep, leftmargin=*,topsep=0pt]
\item Cryptographic Solutions incur excessive overhead, degrading the real-time utility required by the CPHS~\cite{zhang2022homomorphic}.

\item Static Perturbation Solutions  cannot dynamically adjust the privacy level based on a user's instantaneous behavior or the current privacy risk, leading to suboptimal personalized protection~\cite{taherisadr2023adaparl, wang2019privacy}.
 \end{itemize}
\sysname addresses this by using the Early-Exit DRL architecture to learn a dynamic and personalized privacy policy. This allows the system to make a real-time decision on the necessary information processing depth, minimizing the data's descriptive power ($s_t\rightarrow a^*$) only to the point required to maintain sufficient utility.

%% file: 04_privacymetric.tex
\vspace{-1mm}\section{Privacy Metric: Quantifying State Inference}\label{sec:privacymetric}

The fundamental privacy risk is determined by the information flow from the sensitive private state sequence ($\mathcal{S}^{\text{private}}$) to the observed personalized action sequence ($\mathcal{A}^*$). This relationship is entirely governed by the DRL agent's learned \textbf{state-action policy, $\pi$}, which dictates the conditional probability of selecting action $a$ given state $s$: $\pi(a | s)$. The information leakage, representing the adversary's ability to infer $\mathcal{S}^{\text{private}}$ by observing $\mathcal{A}^*$, can be quantified by Mutual Information ($\mathcal{I}$)):
\begin{align}\label{eq:MI}
\mathcal{I}(\mathcal{A}^*;\mathcal{S}^\text{private}) = \sum_{ a^*,\mathbf{s}^{\text{private}}} p(a^*,\mathbf{s}^{\text{private}}) \log \frac{p(a^*, \mathbf{s}^{\text{private}})}{p(a^*)p(\mathbf{s}^{\text{private}}) }
\end{align}

Since $\mathcal{A}^*$ is generated directly by the policy $\pi$ operating on $\mathcal{S}$, the leakage $\mathcal{I}$ is a function of the learned policy. 
This value establishes a \textbf{theoretical upper bound} on the adversary's inferential power~\cite{ito2022success}. If $\mathcal{I}$ is low, no adversary, even one with unlimited computation power, can achieve high prediction accuracy. Conversely, high $\mathcal{I}$ signifies high privacy risk.

Thus, we consider the MI measure $\mathcal{I}_{\pi}(a_t; s_t)$ for a given policy $\pi(s, a)$ as a \textbf{constraint} on the ability to extract information about $s_t$ by observing $a_t$. The total information available to the cloud is, therefore, bounded by this information-theoretic constraint, reinforcing the need for $\pi$ to be privacy-aware.

While MI is an \textbf{application-agnostic} approach and provides a quantifiable measure of information leakage \cite{ito2022success}, it is inherently an average measure and may not fully capture the complexity of privacy decisions in specific contexts. For instance, MI can be sensitive to \textbf{noise in the data}; if the time series data is noisy, MI might not accurately reflect the true correlation between the series. In our setup, however, the edge's local DQN model mitigates some noise effects by generalizing inference effectively before generating $a^*$. Another concern is that MI may be unsuitable for high-dimensional data, suffering from the \textbf{``curse of dimensionality,''} which makes accurate estimation challenging without large datasets. Crucially, since we focus on human-centric CPS applications with a \textbf{finite and limited number of sensitive states and actions}, estimating MI in this context is tractable.

This MI bound is conditional on the policy remaining non-informative at the selected exit; in the discrete, finite state-action spaces of applications, shallower layers produce coarser Q-value estimates that empirically encode less private state information. This will be later confirmed by the MI-vs-exit depth curves in our evaluation section specifically~\ref{fig:pmvmihuman1} and~\ref{fig:miapp2}.

%% file: 05_algorithm.tex
\vspace{-1mm}\section{\sysname Framework}\label{sec:alg}

\sysname framework is our dynamic defense mechanism, architected on the CPHS Edge layer to directly mitigate the inference threat described in Section~\ref{sec:threatmodel}. 

\subsection{CPHS Problem Formulation as an MDP}
We model the personalized control task for each user $i$ in the human-centric CPHS as a Markov Decision Process (MDP), a common approach for handling human variability in these systems \cite{elmalaki2018sentio, elmalaki2022maconauto}, defined by the tuple ($\mathcal{S},\mathcal{A},T,R_i,\gamma$). The State ($\mathbf{s}_t$) is the complete observation at the edge, critically including the sensitive private human state that the cloud adversary attempts to infer. The Action ($a_t$) is the control signal sent to the cloud (e.g., HVAC setpoints), which constitutes the exposed action sequence $\mathcal{A}^*$. $T$ is the transition probability, and $\gamma$ is the discount factor. The DRL agent's objective is to learn the optimal personalized policy $\pi_i$ that maximizes the expected cumulative return based on the Reward ($R_i$), which is task specific: $E_{\pi_i}[\sum_{t=0}^\infty \gamma R_i(\mathbf{s}_t,a_t)]$.








\vspace{-1mm}
\subsection{\sysname Architecture and Dynamic Control}
To achieve the dynamic and personalized tradeoff required by the problem, we employ a modified Deep Q-Network (DQN) architecture featuring an Early-Exit (EE) mechanism. This mechanism acts as the dynamic control knob that modulates the information content of the output action $a^*$ in real-time.

The \sysname agent consists of a single deep neural network backbone with $L$ total layers. We embed a set of $K$ intermediate exit branches $(E_1,E_2,\dots,E_K)$ along the backbone, where $E_k$ is situated after layer $L_k$. The final layer $L$ serves as the last exit $E_{K+1}$. Each exit $E_k$ outputs an action-value estimate: $Q_k(s,a)$.
\begin{itemize}[noitemsep, leftmargin=*,topsep=0pt]
\item \textbf{Decision Mechanism:} At time $t$, the state $s_t$ passes through the network backbone. At each exit $E_k$, a confidence metric $\text{Conf}_k(s_t)$ is calculated. The agent selects the earliest exit $E_{k^*}$ that satisfies the personalized threshold $\tau_i$ defined for utility $u_i$ and privacy $p_i$ budgets:
$k^*=\min\{k\in{1,\dots,K+1}|\text{Conf}_k(s_t)>\tau_i\}$. 
The final action $a_t^*$ is the greedy action derived from the Q-values of the selected exit $E_{k^*}: a_t^*=\arg\max_a Q_{k^*}(s_t,a)$.

\item \textbf{Privacy-Utility Relationship:} The index $k^*$ of the chosen exit directly controls the information flow. An earlier exit ($k^*$ is small) means fewer features are processed, resulting in a less descriptive action $a^*$ and a lower theoretical MI/higher privacy. A later exit offers higher confidence and potentially higher utility, but it comes with a corresponding higher privacy risk (higher MI).
\end{itemize}


This design choice allows \sysname to share a single backbone across all exits, significantly reducing storage and retraining cost. This motivates \sysname's structural approach over multi-policy alternatives where multi-policy gating without EE would require training and storing $K$ complete policies per user.


\begin{figure}[!t]
\centering
\includegraphics[scale=0.5]{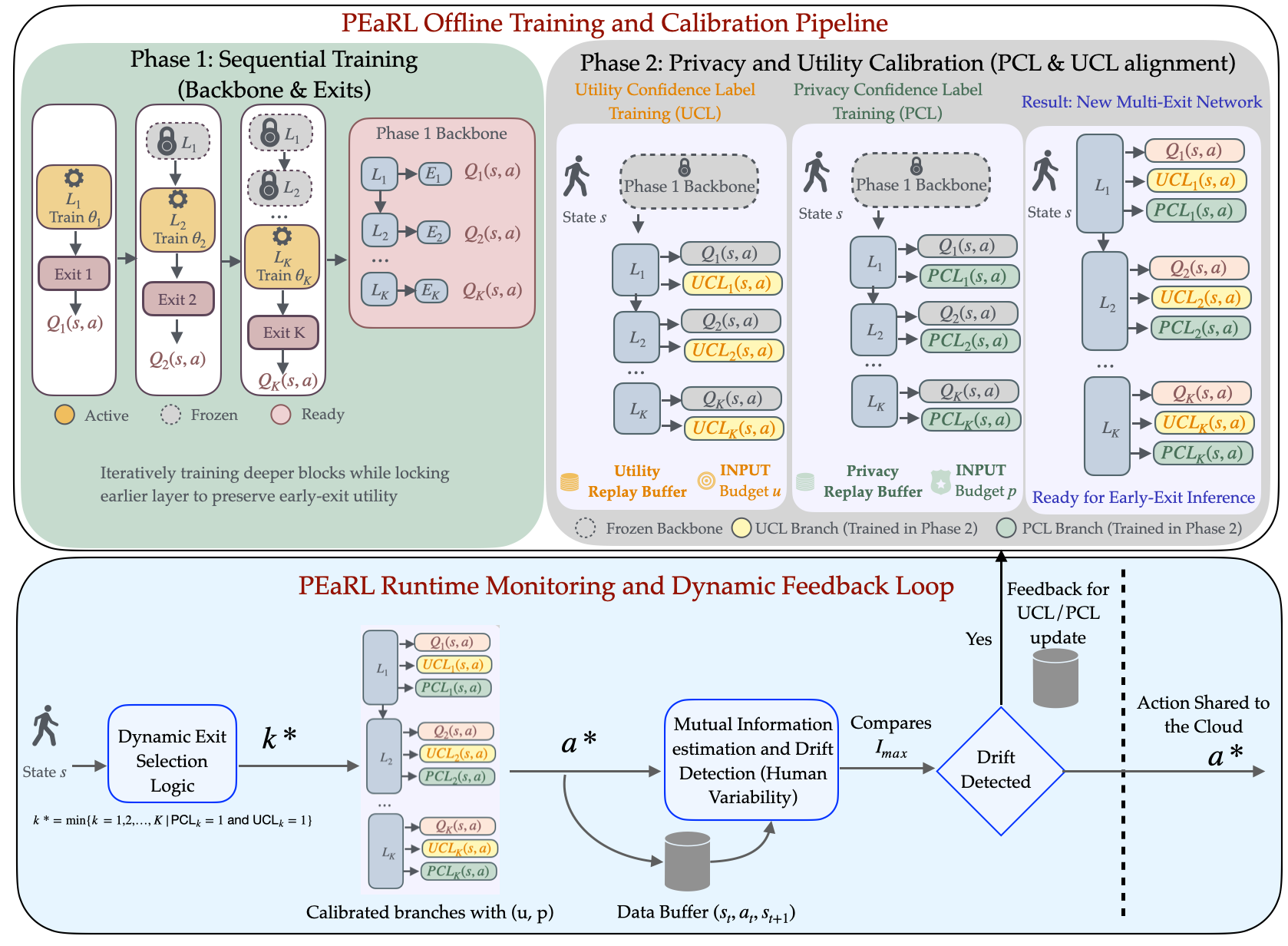}
\caption{\sysname offline training and calibration pipeline (upper panel) and the runtime monitoring and dynamic feedback loop (lower panel). (1) \sysname training includes two phases.  PHASE 1 of the training flow of the \sysname algorithm detailed the sequential training. An exit branch follows each layer. The layer and associated branch are trained in each step, and the next layer and its branch are added in the following step. Training of the new layer and its branch happens while the parameters of the previous layers and branches are frozen. PHASE 2 of the training flow of the \sysname algorithm includes the privacy and utility confidence path (PCL and UCL) training for each exit branch of the DQN. The network has two separate replay buffers. The privacy buffer contains the action, state, and privacy labels for all branches. The utility buffer includes the action, state, and utility labels of the branches. (2) \sysname runtime monitoring and feedback uses the trained multi-exit network to output the action $a$. Drift detection due to human variability is estimated and a feedback signal is sent back to update the confidence training paths (PCL and UCL).}
\label{fig:completeframework}
\end{figure}

\vspace{-1mm}
\subsection{Training \sysname for Personalized Budget Alignment} \label{sec:training}

\sysname's training ensures that each EE branch is aware of the personalized privacy-utility tradeoff before deployment. The training is structured into two main phases:

\vspace{-1mm}
\paragraph{\textbf{Phase 1: Foundational EE-DQN Training}} 
Phase 1 focuses on developing a foundational DQN with integrated EE branches. This phase employs sequential training where the network is gradually expanded. The process begins with initializing a single-layer Q-network. New layers with associated EE branches are introduced and trained one at a time, while keeping previously trained parameters fixed \cite{kosta2022rapid}. This ensures that adding new layers does not affect the Q-value utility performance of existing, shallower branches as illustrated in~\autoref{fig:completeframework}.
\vspace{-1mm}
\paragraph{\textbf{Phase 2: Budget-Constrained Confidence Alignment}}
Phase 2 focuses on training confidence paths that enable the agent to select the optimal EE branch at inference time based on pre-defined budgets. Confidence labels are assigned to each EE branch based on personalized utility ($u$) and privacy ($p$) budgets, with a label of ``1'' indicating that the budget is satisfied as illustrated in Figure~\ref{fig:completeframework}. Training of these labels are done as follows: 

\begin{itemize}[noitemsep, leftmargin=*,topsep=0pt]
\item \textbf{Utility Confidence Label (UCL) Training:} 
UCL training ensures that the action chosen at exit $E_i$ is close to the action that would be chosen by the most accurate branch ($Q_{\max}$). The UCL acts as a \textbf{tolerance check for task performance}. A utility replay buffer stores state, action, and utility labels for B branches. The UCL is assigned based on the maximum Q-value in branch $i$, $Q_{i_{\max}} (s,a)$, relative to the utility budget $u\in(0,1]$ and the overall maximum $Q_{\max}(s,a)$:
 \begin{equation}\label{eq:ucl}
    UCL_{i}(s, a) = \begin{cases} 
        1, & \text{if } Q_{i_{max}}(s, a) > u \cdot  Q_{max}(s, a) \\
        0, & \text{otherwise.}
    \end{cases}
\end{equation}
Here, the budget $u$ defines the acceptable utility performance degradation. If $u=0.9$, the branch $E_i$ must guarantee a Q-value that is at least $90\%$ of the maximum possible Q-value for that state $s$. 

\item \textbf{Privacy Confidence Label (PCL) Training:}
PCL training directly links the EE branch to the information leakage metric $\mathcal{I}(\mathcal{S}^\text{private};\mathcal{A}^*)$ (Section~\ref{sec:privacymetric}). The PCL acts as a \textbf{risk ceiling check on the personalized privacy requirement}. A privacy replay buffer is maintained, updating for every $n$ interactions (as MI requires a time-series). PCL is assigned based on the MI value $\mathcal{I}_i(s;a)$ of branch $i$, relative to the privacy budget $p\in(0,1]$ and the maximum MI observed across all branches, $\mathcal{I}_{\max}(s;a)$:
\begin{equation}\label{eq:pcl}
    PCL_i(\mathbf{s},a) = \begin{cases} 
        1, & \text{if }  \mathcal{I}_i(a;s) < p \cdot \mathcal{I}_{max}(a;s) \\
        0, & \text{otherwise.}
    \end{cases}
\end{equation}
The intuition is that $\mathcal{I}_{\max}$ represents the worst-case leakage. By setting a threshold based on $p$, the agent ensures the information leakage is below a user-defined fraction of this maximum. For example, if $p=0.5$, the branch $E_i$ must ensure its measured information leakage is less than half of the worst-case leakage. This directly translates the abstract concept of MI into a concrete, personalized constraint for the dynamic exit decision.

\end{itemize}

The trained Q-network is augmented with separate utility and privacy paths, outputting the binary UCL and PCL using a binary cross-entropy loss $\mathcal{L}_{\text{BCE}}$. The confidence-label training framework allows each EE branch to achieve a tailored tradeoff based on specific privacy and utility budgets. The two training phases are detailed in~\autoref{alg:algorithms} and depicted in~\autoref{fig:completeframework}. 

We compute the per-step MI $I(a_t; s_t)$ with window $n$ as a practical proxy for sequence-level leakage $I(A_{1:T}; S_{1:T})$. While the full sequence MI is the theoretically correct measure of the cloud's inferential capability, per-step MI is computationally tractable and serves as a conservative lower bound in stationary regimes. The stationarity of the deployed policy ensures that accumulated sequence-level leakage grows approximately proportionally to the per-step estimate, which we acknowledge is an approximation that holds more tightly in the discrete, low-dimensional settings of this work

In practice, MI is estimated using a histogram-based (binning) approach over a sliding window of $n$ interaction steps. Given the finite and discrete nature of both state and action spaces in our considered CHPS applications,
the joint and marginal distributions described in Equation~\ref{eq:MI} are estimated as empirical frequencies over the window, making this approach tractable and consistent~\cite{ito2022success}. The window length $n$ is set as an application-dependent parameter. Bias from finite samples is inherent in histogram estimators but is mitigated by the bounded, low-cardinality spaces.


\paragraph{\textbf{Design Choice}} A related body of work addresses MI-constrained decision-making in dynamical systems, where MI between private states and shared observations is minimized as a regularization term in the policy optimization objective. These approaches, including information-bottleneck-inspired RL and MI-constrained belief-MDP frameworks, derive computable MI upper bounds and integrate them into policy gradient or value-based methods~\cite{taherisadr2023adaparl}. \sysname differs in a key architectural respect: rather than incorporating MI as a penalty during training (which modifies the reward signal and can compromise control stability in physical systems), \sysname enforces privacy structurally at inference time by selecting the shallowest exit that satisfies user-defined budgets. This separation ensures that the backbone DQN optimizes purely for task utility during \textbf{Phase 1}, while \textbf{Phase 2} overlays privacy enforcement without altering the Q-values. Additionally, \sysname's personalized budget system ($u$, $p$) allows per-user privacy calibration, which static MI-regularized policies do not support. The decision to use structural early-exit control rather than policy regularization is also motivated by the CPS setting, where any noise injected into control signals propagates through the physical plant.


\vspace{-1mm}
\subsection{\sysname Personalized Inference and Adaptation}
\paragraph{\textbf{Personalized Exit Selection}}
Upon deployment, the DQN, now equipped with confidence paths, makes real-time decisions. The selection of the early exit ($E_k^*$) is determined by the specific personalized privacy ($p$) and utility ($u$) budgets: the agent chooses the earliest branch that satisfies both the UCL and PCL requirements for the current state. 

In the event of a Budget Failure (i.e., when no exit branch satisfies both the personalized Utility Confidence Label ($UCL=1$) and Privacy Confidence Label ($PCL=1$)), the system prioritizes System Integrity (Utility). The agent defaults to selecting the earliest exit branch that satisfies $UCL=1$. This ensures the resulting action $a^*$ is near-optimal for the task, minimizing the loss of system functionality, even if it slightly compromises the user's preferred privacy budget ($p$). In practice, fallback is most likely under high utility demand ($u$ close to 1), where the utility requirement alone forces deeper exits that inherently carry higher MI. To mitigate the privacy implications of fallbacks, practitioners should select budget pairs $(u, p)$ that have at least one compliant exit. The system design intentionally prioritizes utility in fallback to prevent control instability in safety-critical CPS, a tradeoff we discuss in Section~\ref{sec:discuss}.

The inference is detailed in Algorithm~\ref{alg:exit_branches_inference}.

 \vspace{-2mm}
\paragraph{\textbf{Incorporating Human Variability}}

As human behavior and preferences are not static, the MI profile of the policy $\pi$ can change over time. \sysname continuously monitors the real-time MI variation from the deployed policy. If this variation drops below a predetermined threshold, $I_\text{threshold}=v\cdot I_{\max}$ (where $v\in(0,1]$ and $\mathcal{I}_{\max}$ is the maximum MI at convergence), it indicates a significant change in human behavior or the environment. When \sysname received this feedback signal, it initiates a full retraining phase (both Phase 1 and Phase 2) to adapt the EE-DQN and its confidence paths to the evolving user profile, ensuring the personalized tradeoff remains optimal over the long term. This is detailed in Algorithm~\ref{alg:inference_with_retraining}.

A key differentiator of \sysname from information-regularized baselines is that privacy enforcement happens at inference time via architectural structure rather than during policy optimization. This means the underlying Q-learning is unperturbed, preserving control stability, while privacy is enforced post-hoc through exit selection.

\begin{algorithm}[!t]
\small
\caption{PEARL Training: Foundational DQN and Confidence Alignment}
\label{alg:algorithms}
\begin{algorithmic}[1]
\State \textbf{Input:} Environment $\mathcal{E}$, Number of EE Branches $B$, Utility Budget $u$, Privacy Budget $p$, Privacy Monitoring Period $n$. 

\State \textbf{Output:} Trained \sysname DQN Q with utility and privacy confidence labels.

\State \textbf{Initialize:} DQN $Q$ with $B$ exits, 
Utility Replay Buffer $\mathcal{R}_{\text{Util}}$, Privacy Replay Buffer $\mathcal{R}_{\text{Priv}}$.

\State \textbf{PHASE 1: Foundational EE-DQN Training (Sequential)}

\For{$i = 1$ to $B$}
    \State Train Layer $L_i$ and Exit $E_i$ sequentially while freezing $L_{j<i}$.
    \State Execute DQN steps to train $Q_i(\mathbf{s}, a)$, maximizing task reward.
\EndFor
\State After Convergence:
\State Record $\mathcal{I}_{\max} \leftarrow \max (\mathcal{I}_i(\mathbf{s}; a)) \forall i \in B$. 
\State  
Record $Q_{i_{\max}} \leftarrow \max_{a} Q_i(\mathbf{s}, a) \forall i \in B$.        
\State Record $Q_{max}(\mathbf{s}, a)\leftarrow \max (Q_{i_{\max}}(\mathbf{s}; a)) \forall i \in B$. 

\State \textbf{PHASE 2: Budget-Constrained Confidence Path Training}
\State Given a trained DQN $Q$ with $B$ branches. 
\State Initialize $f_{\text{UCL}_i}$ and $f_{\text{PCL}_i}$ (Confidence Paths) $\forall i \in B$.
\For{each training step $t$}
    \State Interact $\mathcal{E}$ by observing the $\mathbf{s_t}$ and receiving $a_i$ from each branch $i$ in Q.
    
    \State Choose one $a_i$ randomly from $B$. 
    
    \State \textbf{Calculate Confidence Labels:}
    \For{$i = 1$ to $B$}
        \State $UCL_i \leftarrow \mathbf{1}[Q_{i_{\max}} > u \cdot Q_{\max}]$  \Comment{Utility Eq.~\ref{eq:ucl}}
        \State Update $\mathcal{R}_{\text{Util}}$ with $UCL_i$:  \State $\mathcal{R}_{\text{Util}} = \{\mathbf{s}_t,a, UCL_1,\dots,UCL_B \}$  
        \If{$t \pmod n = 0$}
            \State $\mathcal{I}_i \leftarrow \text{Compute MI}(a^*, \mathbf{s})$ for $E_i$ over last $n$ steps.
            \State $PCL_i \leftarrow \mathbf{1}[\mathcal{I}_i < p \cdot \mathcal{I}_{\max}]$ \Comment{Privacy Eq.~\ref{eq:pcl}}
            \State Update $\mathcal{R}_{\text{Priv}}$ with $PCL_i$: 
            \State $\mathcal{R}_{\text{Priv}} = \{\mathbf{s}_t,a, PCL_1,\dots,PCL_B \}$
        \EndIf
    \EndFor
  \EndFor  
    \State \textbf{Train Confidence Paths:}
    \For{each training step $t$}
        \State Sample from  $\mathcal{R}_{\text{Util}}$ and $\mathcal{R}_{\text{Priv}}$:
        \State Minimize $\mathcal{L}_{\text{BCE}}(f_{\text{UCL}}(\mathbf{s}_t), UCL_i) \forall i \in B$.
        \State Minimize $\mathcal{L}_{\text{BCE}}(f_{\text{PCL}}(\mathbf{s}_t), PCL_i) \forall i \in B$.
    \EndFor
\end{algorithmic}
\end{algorithm}

\begin{flushleft}
\begin{algorithm}
\small
\caption{PEARL Dynamic Exit Selection (Inference)}
\label{alg:exit_branches_inference}
\begin{algorithmic}[1]

\State \textbf{Input:} Current State $\mathbf{s}_t$, Trained \sysname DQN $Q$ (with $f_{\text{UCL}}, f_{\text{PCL}}$) 
\State \textbf{Output:} Selected Action $a_t^*$.
\State $k_{\text{selected}} \leftarrow \text{None}$
\State $k_{\text{fallback}} \leftarrow \text{None}$ \Comment{Initialize fallback to highest-utility option}
\For{$k = 1$ to $B$} \Comment{Iterate through exit branches from shallowest to deepest}
    \State $\text{Confidence}_{\text{Util}} \leftarrow f_{\text{UCL}}(\mathbf{s}_t)[k]$ \Comment{Output $\in \{0, 1\}$}
    \State $\text{Confidence}_{\text{Priv}} \leftarrow f_{\text{PCL}}(\mathbf{s}_t)[k]$ \Comment{Output $\in \{0, 1\}$}
    
    \If{$\text{Confidence}_{\text{Util}} = 1$ \textbf{and} $\text{Confidence}_{\text{Priv}} = 1$}
        \State $k_{\text{selected}} \leftarrow k$ \Comment{Success: Select the earliest branch satisfying both budgets}
        \State \textbf{break}
    \EndIf
    \If{$k_{\text{fallback}} = \text{None}$ \textbf{and} $\text{Conf}_{\text{Util}} = 1$}
        \State $k_{\text{fallback}} \leftarrow k$ \Comment{Track the earliest high-utility option as fallback}
    \EndIf
\EndFor
\If{$k_{\text{selected}} \neq \text{None}$}
    \State $k^* \leftarrow k_{\text{selected}}$
\Else
    \State $k^* \leftarrow k_{\text{fallback}}$ \Comment{Budget Failure: Default to the earliest high-utility action}
\EndIf
\State $a_t^* \leftarrow \arg\max_a Q_{k^*}(\mathbf{s}_t, a)$ 
\State \textbf{return} $a_t^*$
\end{algorithmic}
\end{algorithm}
\end{flushleft}


\begin{algorithm}
\small
\caption{PEARL Inference with Continuous Adaptation to Human Variability}
\label{alg:inference_with_retraining}
\begin{algorithmic}[1]
\State \textbf{Input:} Trained PEARL $Q$, MI Threshold Factor $v$, MI Monitoring Period $n$, Maximum MI $\mathcal{I}_{\max}$.

\State \textbf{Initialize:} $\mathcal{I}_{\text{hist}} \leftarrow []$, $\mathcal{I}_{\text{threshold}} \leftarrow v \cdot \mathcal{I}_{\max}$.

\Loop \Comment{Continuous Inference}
    \State Observe $\mathbf{s}_t$.
    \State $a_t^* \leftarrow \text{Algorithm~\ref{alg:exit_branches_inference}}(\mathbf{s}_t, Q)$ \Comment{Perform dynamic exit selection}
    \State Execute $a_t^*$, observe $\mathbf{s}_{t+1}$.
    
    \State Store interaction $(\mathbf{s}_t, a_t^*, \mathbf{s}_{t+1})$ in monitoring buffer $\mathcal{M}$.
    
    \If{buffer $\mathcal{M}$ size reaches $n$}
        \State $\mathcal{I}_{\text{current}} \leftarrow \text{Compute MI}(\mathcal{M}_{\mathcal{A}^*}, \mathcal{M}_{\mathcal{S}^{\text{private}}})$
        \State $\mathcal{I}_{\text{hist}} \leftarrow \mathcal{I}_{\text{hist}} \cup \{\mathcal{I}_{\text{current}}\}$
        
        \Comment{Check for significant behavioral change (MI drop)}
        \If{$|\mathcal{I}_{\max} - \mathcal{I}_{\text{current}}| > \mathcal{I}_{\text{threshold}}$}
            \State $\mathcal{I}_{\max} \leftarrow \mathcal{I}_{\text{current}}$ \Comment{Update $\mathcal{I}_{\max}$ to new baseline}
            \State \textbf{Retrain:} $\text{Algorithm~\ref{alg:algorithms}}(\mathcal{E}, B, u, p, n)$
        \EndIf
        
        \State Clear monitoring buffer $\mathcal{M}$.
    \EndIf
\EndLoop
\end{algorithmic}
\end{algorithm}

%% file: 06_application1_home.tex
\vspace{-2mm}
\section{Application~1: Thermal House Model}\label{sec:app1house}

Recent advancements in smart heating, ventilation, and air conditioning (HVAC) systems integrate human input to improve user satisfaction~\cite{jung2017towards, elmalaki2021fair}. 
For example, body temperature decreases during sleep~\cite{barrett1993sleep} and increases during physical exercise~\cite{lim2008human}. 
Edge devices can monitor these human conditions, such as sleep patterns and physical activities, by gathering relevant sensor data~\cite{nguyen2016lightweight,shany2012sensors}.
This section evaluates \sysname framework within HVAC system, where the sensitive private state is the user's activity. The evaluation demonstrates \sysname's capability to provide a personalized, adaptive privacy-utility tradeoff by dynamically controlling the information content of the shared HVAC actions.

\vspace{-2mm}
\subsection{Human-Centric System Design}

Our study simulates a thermodynamic model of residential houses. 
The heating system uses a heater emitting air at $50^{\circ}C$, while cooling is achieved with a cooler emitting air at $10^{\circ}C$. A thermostat maintains the indoor temperature, allowing a deviation of up to $2^{\circ}C$ above or below the setpoint, controlled externally by \sysname~\cite{MATLABther}. Humans are modeled as thermal sources, with heat generation influenced by average exhale breath temperature (EBT) and respiratory minute volume (RMV)\cite{elmalaki2021fair,carroll2006elsevier}.

\begin{figure*}[!t]
\centering
\begin{minipage}{0.45\textwidth}
    \centering
\includegraphics[width=\linewidth]{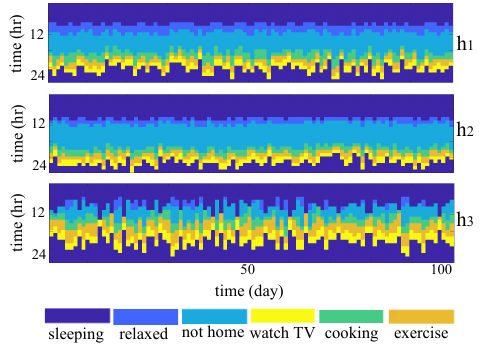}\vspace{-1mm}
\caption{Human activity profile for three humans.\vspace{-3mm}}
\label{fig:actall}
\end{minipage}
\begin{minipage}{0.45\textwidth}
\centering
\includegraphics[width=\linewidth]{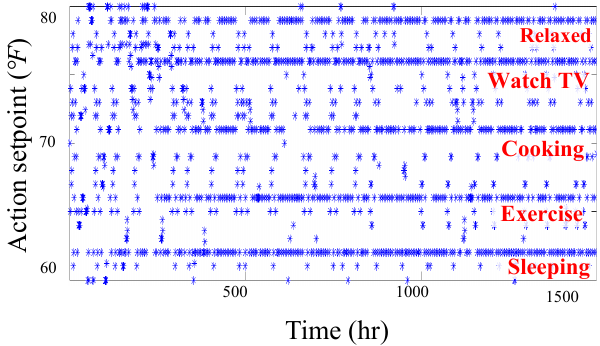}
\caption{Action convergence of human $\mathbf{H_1}$  using $\mathbf{DQN}$.\vspace{-1mm}}
\label{fig:actconvall}
\end{minipage}
\end{figure*}

We simulate the behavior of three distinct individuals in three homes by varying their RMV and metabolic rates. Our simulation includes six activity categories: sleeping, relaxing, watching TV, cooking, exercising, and a state called ``not at home.'' Each individual follows unique lifestyle patterns. Human $H_1$	maintains a structured weekly routine with minimal randomness, 
$H_3$ has an irregular lifestyle with frequent unexpected changes, and $H_2$ exhibits moderate randomness between $H_1$ and $H_3.$ ~\autoref{fig:actall} illustrates the activity patterns of 
$H_1$, $H_2$, and $H_3$.

To introduce variability, activities are randomly assigned within the same time slots, creating diverse daily behaviors. Our model does not account for age, sex, or time of day. We expanded the thermal house model provided by MathWorks to include a cooling system and our human model~\cite{MATLABther}. Although more complex simulators like EnergyPlus exist for comprehensive energy consumption assessments in smart buildings~\cite{gerber2014energyplus}, we use a simplified thermal house model to evaluate \sysname. Each simulation step represents one hour, with 24 steps equivalent to one day.

\vspace{-2mm}
\subsection{DRL Design}

The control task is formalized as an MDP where the State ($\mathcal{S}$) includes the human activity and indoor temperature: $\mathcal{S} = \{(act, temp): act \in [1, 6] \text{ and } temp \in [60, 80]\}$. the Action ($\mathcal{A}$) is the discrete temperature setpoint shared with the cloud. The Reward ($\mathcal{R}$) is based on the Predicted Mean Vote (PMV) to maximize thermal comfort (i.e., maintaining PMV within $[-0.5,0.5]$~\cite{fanger1970thermal}). In practical applications, PMV can be estimated using edge devices, such as black globe thermometers~\cite{Globe}.



\vspace{-2mm}
\subsection{\sysname Adaptation \& Evaluation}

Our evaluation strategy begins by demonstrating \sysname's capability to adapt to various human behavior patterns by selecting different early-exits for each human to maximize utility. (\autoref{sec:humanadapt}). We then assess the privacy leakage resulting from this personalization based on our threat model (\autoref{sec:privacyleakapp1}). Next, we examine the impact of employing \sysname with $PCL$ and $UCL$ at specific privacy $p$ and utility $u$ budgets on mitigating privacy leaks (\autoref{sec:privacyadaptationapp1}). Finally, we explore the trade-off between choosing $p$ and $u$ in this application. 
Our results indicate that \sysname effectively provides personalized privacy-aware DQN tailored to human behavior.

\subsubsection{\textbf{Architectural Baseline}}\label{sec:humanadapt}

\begin{figure*}[!t]
\centering
\includegraphics[width=0.995\textwidth]{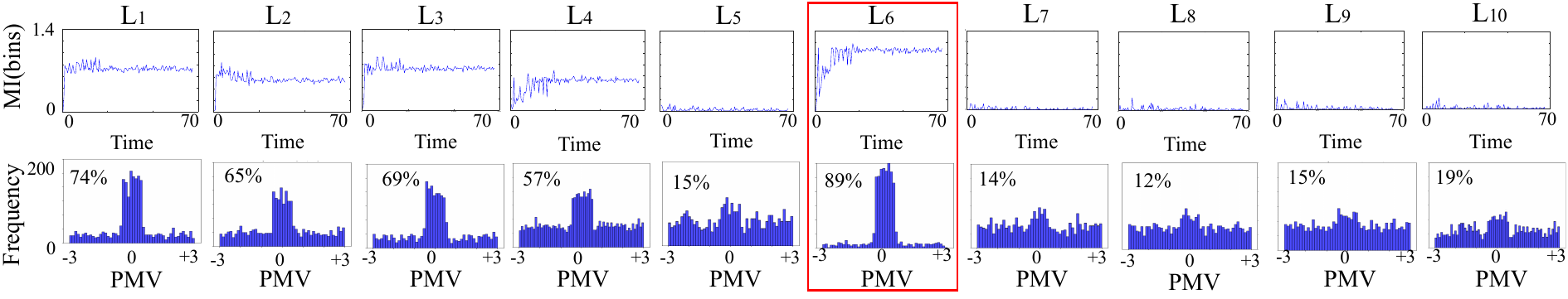}
\caption{PMV and corresponding MI at every branch for human $\mathbf{H_1}$. Top: mutual information between the current state and the chosen action in the associated exit layer. Bottom: PMV histogram for the associated exit with the percentage of the PMV values in the comfortable range ($\mathbf{[-0.5,+0.5]}$). Mutual information and PMV percentage in the comfortable range occurs at the layer $\mathbf{6}$ exit branch.}
\label{fig:pmvmihuman1}
\end{figure*}

To establish a baseline, we trained 10 Deep Q-Network (DQN) without exit branches, varying the number of layers from $1$ to $10$. We evaluated the performance of these models based on their ability to maintain the thermal comfort PMV $[-0.5, +0.5]$. Our analysis revealed that optimal performance varied across individuals, with different layer depths yielding the best results. For instance, a 6-layer DQN was optimal for $H_1$, while $8$ and $3$ layers were best for $H_2$ and $H_3$, respectively. These findings suggested that different DQN depth is tailored to different human behaviors. Using this insight, we decide on a $10$-layer base DQN to provide sufficient capacity for \sysname's architectural design.~\autoref{fig:actconvall} illustrates the convergence of actions during training for a 6-layer DQN, showing how the model converges to specific actions for human $H_1$.

Accordingly, we designed a $10$-layer DQN with $10$ EE-branches, each consisting of $2$ fully connected layers for each human. This EE-DQN was trained using \textbf{PHASE 1} in \sysname, which involves ``sequential training'' with up to $10$ layers. Three independent DQNs with EE were trained for the three individuals\footnote{A batch size of $16$ (1 sample per hour) was used to train the model, which means it takes $16$ hours to have enough samples in the replay to update the model parameters.}. 

\begin{itemize}[noitemsep, topsep=0pt, leftmargin=*]
\item \textbf{Initial Baseline:} Analysis showed that the utility-optimal performance (maximizing the percentage of PMV samples in the comfortable range) occurred at different depths for each user (e.g., Layer 6 for $H_1$, Layer 8 for $H_2$). The second row in \autoref{fig:pmvmihuman1} presents a histogram of PMV values with the percentage indicating the proportion of PMV values within the comfortable range $[-0.5,+0.5]$. \textbf{This confirms that DQN depth must be personalized for optimal utility.}
\item \textbf{Utility-Privacy Correlation:} For $H_1$, the peak PMV percentage occurred at Branch 6, which also coincided with the peak Mutual Information (MI) (Figure~\ref{fig:pmvmihuman1}, top row). This established the fundamental conflict: \textbf{higher utility necessitates deeper processing, leading to increased information leakage (MI).}
\end{itemize}


\subsubsection{\textbf{Privacy Leak Assessment}}\label{sec:privacyleakapp1}

Following the threat model in \autoref{sec:threatmodel}, we simulated an ``honest-but-curious'' cloud monitoring the exposed action sequence ($\mathcal{A}^*$). 

\begin{figure}[!t]
\centering
\includegraphics[width=0.5\linewidth]{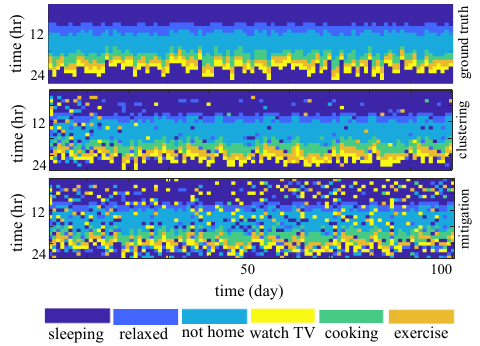}\\
\caption{Activity pattern of the house model. Top: ground truth of the activity patterns of human $\mathbf{H_1}$. Middle: clustering result (using $K$-means) as inferred by the cloud with $\mathbf{87\%}$ accuracy. Bottom: clustering result (using $K$-means) with \sysname at $\mathbf{p=0.7}$ and $\mathbf{u=0.75}$.}
\label{fig:clustermitigact}
\end{figure}

\begin{itemize}[noitemsep, topsep=0pt, leftmargin=*]
\item \textbf{``Honest-but-Curious'' Inference:} The cloud applied unsupervised clustering ($K$-means) to the HVAC setpoints shared by the utility-optimal branch ($L_6$ for $H_1$). The elbow method on the Within-Cluster Sum of Squares (WCSS) against the number of clusters 
correctly identified six clusters (without prior knowledge), corresponding to the six actual human activities.

\item \textbf{Leakage Quantification:} Comparing the clustered patterns with the ground truth activity (Figure \ref{fig:clustermitigact}, top/middle), the cloud achieved 87\% accuracy in inferring $H_1$'s behavioral patterns using 100 simulated days. Although the cloud does not infer specific activities, it can deduce behavioral patterns using its domain knowledge by making educated guesses about when the individual likely sleeps, leaves, or deviates from their routine. This confirms that, without mitigation, the exposed control actions leak sensitive information, validating the need for the \sysname framework.
\end{itemize}

The K-means clustering adversary used in this evaluation represents a lower-bound attack scenario consistent with an honest-but-curious cloud exploiting domain knowledge without training labeled classifiers. While more capable sequence-level adversaries (e.g., HMMs or RNNs) could potentially achieve higher baseline inference accuracy, the information-theoretic argument in Section~\ref{sec:privacymetric} establishes that reducing MI places a fundamental ceiling on any adversary's performance regardless of computational sophistication. The clustering result thus serves as a conservative, interpretable validation consistent with the honest-but-curious threat model.

Crucially, the MI-based defense is adversary-agnostic: since MI $I(\mathcal{A}^*; \mathcal{S}^{\text{private}})$ constitutes a theoretical upper bound on any adversary's inference accuracy (Equation~\ref{eq:MI}), a reduction in MI directly implies reduced leakage against all adversaries, including more sophisticated temporal models (HMMs, RNNs). The clustering evaluation empirically validates this bound in the specific application context.

We acknowledge that evaluating against stronger adversaries (sequential classifiers, temporal models) would further validate the defense. However, since our privacy metric is MI-based and MI is an information-theoretic bound on any adversary's performance, the clustering results serve as an empirical sanity check rather than an adversary-specific evaluation.




\begin{figure}[!t]
\begin{minipage}{\columnwidth}
\captionof{table}{$\mathbf{PCL}$ and $\mathbf{UCL}$ confidence per layer for all humans at various values of $\mathbf{u}$ and $\mathbf{p}$. }
    \label{table:allexitalow}
\centering

\resizebox{\textwidth}{!}{
    \begin{tabular}
   {|c||c|c|c||c|c|c||c|c|c||c|c|c||c|c|c||}
        \hline
         & \multicolumn{3}{c||}{\cellcolor[HTML]{ADD8E6}u = 0.95} & \multicolumn{3}{c||}{\cellcolor[HTML]{ADD8E6}u = 0.85} & \multicolumn{3}{c||}{\cellcolor[HTML]{ADD8E6}u = 0.75}   & \multicolumn{3}{c||}{\cellcolor[HTML]{ADD8E6}u = 0.65}  & \multicolumn{3}{c||}{\cellcolor[HTML]{ADD8E6}u = 0.55}  \\
        \cline{1-16}
         Human & $H_1$ & $H_2$ & $H_3$  & $H_1$ & $H_2$ & $H_3$  &  $H_1$ & $H_2$ & $H_3$  &  $H_1$ & $H_2$ & $H_3$  &  $H_1$ & $H_2$ & $H_3$  \\
        \hline
        \cellcolor[HTML]{ADD8E6}p = 0.9 & $\times$ & $\times$  & $\times$  & $\times$  & $\times$  & $\times$  & $\times$  & $\times$  & $\times$  & $\times$  & $\times$  &  $L_{3}$  & $L_{4,6}$  & $L_{8}$  &  $L_{2,3}$   \\
        \hline
        
        \cellcolor[HTML]{ADD8E6}p = 0.8 & $\times$  & $\times$  & $\times$  & $\times$  & $\times$  & $\times$  & $L_6$   &  $\times$  & $L_3$ & $L_{1,6}$  & $L_{5,8}$   &  $L_{3,9}$  & $L_{1,3,6}$  & $L_{5,8}$  &  $L_{3,9}$    \\
        \hline
        
        \cellcolor[HTML]{ADD8E6}p = 0.7 & $\times$  & $\times$  & $\times$  & $\times$  & $\times$  & $\times$  & 
        $L_{1,6}$  & $L_{8}$  &  $L_{3,5}$ & 
        $L_{1,4,6}$  & $L_{3,6,9}$   &  $L_{4,7}$  & 
        $L_{1,6}$  & $L_{3,6,8}$  &  $L_{3,5,7}$    \\
        \hline
        \cellcolor[HTML]{ADD8E6}p = 0.6 & $\times$  & $\times$  & $\times$  & $\times$  & $\times$  & $\times$  & $L_{1,6,7}$  & $L_{5,8}$  &  $L_{3,4,5}$ & 
        $L_{1,6,7}$  & $L_{5,8}$   &  $L_{3,4,5}$  & 
        $L_{1,6,7}$  & $L_{4,5,7}$  &  $L_{4,7,8,9}$\\
        \hline
    \end{tabular}
    } 
    
\centering
\end{minipage}
\vspace{-3mm}
\end{figure}


\vspace{-1mm}
\subsubsection{\textbf{\sysname Personalized  Adaptation}}\label{sec:privacyadaptationapp1}

The core evaluation involved running \textbf{PHASE 2} (Confidence Alignment) to train the $UCL$ and $PCL$ paths based on personalized budgets ($u$ and $p$) (Algorithm~\ref{alg:algorithms}) and then applying the Dynamic Exit Selection (Algorithm~\ref{alg:exit_branches_inference}).

We analyzed the impact of budgets on exit selection by varying the budgets $u$ and $p$ to constrain which exit branches are available (Table \ref{table:allexitalow}).

\begin{itemize}[noitemsep, topsep=0pt, leftmargin=*]
\item \textbf{Strict Constraints (High $u$, High $p$):} When demanding high utility (e.g., $u=0.95$), the number of acceptable exits is severely limited, often resulting in no possible exit, as a high utility goal usually violates the privacy constraint\footnote{During inference, we track the earliest high-utility option for fallback.}. 

\item \textbf{Relaxed Constraints (Low $u$, Low $p$):} Relaxing both constraints (e.g., $u=0.65,p=0.6$) increases the available exit branches (e.g., $L_{1,6,7}$ for $H_1$), giving the agent more flexibility to choose a shallower, privacy-preserving path.

\item \textbf{Personalization:} The available exits differ  across users even for the same budget, demonstrating that \sysname successfully tailors the policy to individual behavioral differences (e.g., at $u=0.75,p=0.7$, $H_1$ uses $\{L_1, L_6\}$ while $H_3$ uses $\{L_3$, $L_5\}$).
\end{itemize}

\vspace{-1mm}
\subsubsection{\textbf{Privacy-Utility Tradeoff}}\label{sec:putresultapp1}
\begin{figure*}[!t]
\centering
\begin{minipage}{0.45\textwidth}
    \centering
    \includegraphics[scale=0.6]{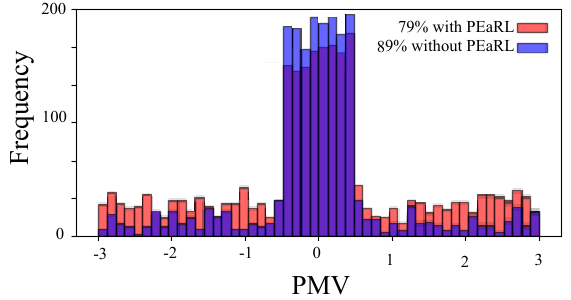}
    \caption{PMV before and after applying \sysname using utility budget $\mathbf{u=0.75}$ and privacy budget $\mathbf{p=0.7}$ for $\mathbf{H_1}$.}
    \label{fig:pmvbeforafter}
\end{minipage}\hfill
\begin{minipage}{0.45\textwidth}
    \centering
    \includestandalone[width=\linewidth]{figures/app1/app1put}
    \caption{Privacy-utility tradeoff for three humans with various values of $\mathbf{u}$ at $\mathbf{p=0.7}$.}
    \label{fig:putallup}
\end{minipage}\hfill

\vspace{-3mm}
\end{figure*}

We analyzed the system's performance using a sample constraint set ($u=0.75,p=0.7$).
\begin{itemize}[noitemsep, topsep=0pt, leftmargin=*]
\item \textbf{Privacy Enhancement:} When PEARL selected the budget-compliant exit, the cloud's clustering accuracy for $H_1$ dropped significantly from 87\% (unmitigated) to 61\% (mitigated) (Figure \ref{fig:clustermitigact}, bottom), representing a 26\% improvement in privacy. Similar improvements were seen for $H_2$(23\%) and $H_3$(21\%).

\item  \textbf{Utility Cost:} This privacy gain came with a controlled reduction in utility. The percentage of PMV samples within the comfortable range for $H_1$ decreased from 89\% (optimal utility) to 79\% (budget-compliant) (Figure \ref{fig:pmvbeforafter}), indicating a 10\% reduction in performance. Similar performance reductions of $12\%$ and $16\%$ were observed for $H_2$ and $H_3$, respectively.

\item  \textbf{Tradeoff Analysis:} The general analysis (Figure \ref{fig:putallup}) confirmed the inverse relationship between the standard deviation (STD) of PMV (utility metric) and clustering accuracy (privacy metric). Lowering the utility requirement (decreasing $u$) significantly improved privacy, achieving an average 34\% reduction in clustering accuracy across all humans.

\end{itemize}

\begin{figure*}[!t]
\centering
\begin{minipage}{0.9\textwidth}
    \centering
    \includegraphics[scale=1.3]{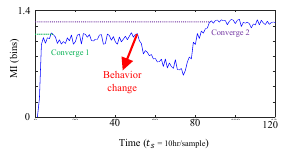}
    \caption{Human behavior changes over time and the corresponding MI at $\mathbf{u=0.75}$ and $\mathbf{p=0.7}$.} 
    \label{fig:variablityofmiadapt}
\end{minipage}
\end{figure*}




\vspace{-1mm}
\subsubsection{\textbf{Adapting to Human Variability}}\label{sec:variability}

Algorithm~\ref{alg:inference_with_retraining} (Continuous Adaptation) was evaluated by simulating a shift in human behavior from $H_3$'s pattern to $H_1$'s pattern after $25$ days.

\begin{itemize}[noitemsep, topsep=0pt, leftmargin=*]
\item \textbf{Monitoring and Retraining:} As the behavior changed, the current Mutual Information ($\mathcal{I}_{\text{current}}$) of the deployed branch decreased, reflecting a divergence from the learned policy's assumptions (\autoref{fig:variablityofmiadapt}). Once $\mathcal{I}_{\text{current}}$
dropped below the threshold ($\mathcal{I}_{\text{threshold}}=0.8$), the system triggered a full retraining (Algorithm 1).

\item \textbf{Adaptation Success:} After approximately $10$ days of retraining, the MI curve stabilized at a new, higher value (Figure \ref{fig:variablityofmiadapt}), confirming that \sysname successfully adapted its policy to the new behavioral pattern, ensuring the privacy-utility tradeoff remained effective in a dynamic environment.
\end{itemize}

In summary, this simulation validates that \sysname successfully achieves personalized and dynamic control over the privacy-utility tradeoff for human-centric HVAC systems, effectively mitigating information leakage while maintaining acceptable thermal comfort, which motivates its evaluation in a real-world context.



%% file: figures/app1/app1put.tex
\begin{tikzpicture}[font=\large\bfseries]
\begin{axis}[
    xlabel={Utility budget},
    ylabel={PMV STD},
    xmin=0.63, xmax=0.87,
    ymin=0.75, ymax=1.7,
    xtick={0.65, 0.75, 0.85},
    ytick={0.8, 1.0, 1.2, 1.4, 1.6},
    legend style={at={(1.2,1)}, anchor= north west},
    y axis line style={-},
    x axis line style={-},
    ylabel near ticks,
    grid=major
    ]

    \addplot[mark=*, blue, dashed] coordinates {(0.65, 1.58) (0.75, 1) (0.85, 0.9)}; 
    \addplot[mark=*, red, dashed] coordinates {(0.65, 1.6) (0.75, 1.4) (0.85, 1.2)}; 
    \addplot[mark=*, Green, dashed] coordinates {(0.65, 1.3) (0.75, 1.3) (0.85, 1.1)}; 
    
    \legend{PMV STD (H1), PMV STD (H2), PMV STD (H3)}
\end{axis}

\begin{axis}[
    axis y line*=right,
    axis x line=none,
    ylabel={Privacy Leak (\%)},
    xmin=0.63, xmax=0.87,
    ymin=0.475, ymax=0.95,
    ytick={0.5, 0.6, 0.7, 0.8, 0.9, 1.0},
    legend style={at={(1.2,0.7)}, anchor= north west},
    y axis line style={-},
    x axis line style={-},
    ylabel near ticks,
    grid=major
    ]
    \addplot[mark=x, blue] coordinates {(0.65, 0.65) (0.75, 0.79) (0.85, 0.89)}; 
    \addplot[mark=x, red] coordinates {(0.65, 0.57) (0.75, 0.6) (0.85, 0.79)}; 
    \addplot[mark=x, Green] coordinates {(0.65, 0.59) (0.75, 0.7) (0.85, 0.8)}; 
    
    \legend{Privacy Leak (H1), Privacy Leak (H2), Privacy Leak (H3)}
\end{axis}

\end{tikzpicture}

%% file: 07_application2_learning.tex
\vspace{-2mm}
\section{Application~2: Smart Classroom with VR}\label{sec:app2learn}

This section presents the real-world deployment of the \sysname framework in a human-centric smart classroom utilizing Virtual Reality (VR) technology. This application is designed to enhance personalized learning by dynamically adjusting the learning environment based on the participant's mental state, while enforcing individual privacy preferences regarding that state. This application is motivated by recent advancements in the education sector, which increasingly emphasizes personalized and remote learning environments. It is well-documented that during extended training or educational sessions, particularly those conducted online or remotely, human performance can decline due to distractions, drowsiness, and fatigue~\cite{terai2020detecting}.

\vspace{-2mm}
\subsection{Experimental Setup}

The experiment engaged 15 participants (aged 20 to 30) using a VR setup to present educational content sourced from Khan Academy (biology~\cite{KhanBiology}, chemistry~\cite{KhanChemistry}, physics~\cite{KhanPhysics}). Our experiment protocol is IRB-approved: ``\#2995 Human Learning Monitoring.'' 

\begin{itemize}[noitemsep, topsep=0pt, leftmargin=*]

\item \textbf{Content Modes:} Educational content was presented in both 2D (conventional laptop screens) and 3D (Oculus VR device) modes. 2D lectures were converted for VR presentation.

\item \textbf{Data Collection:} Participants wore the EMOTIV EPOC$\_$X 14-channel portable EEG device (Figure \ref{fig:emotiv}) to measure mental states, which are indicative of learning ability and cognitive performance~\cite{eslinger1985severe}.

\item \textbf{Procedure:} Each lecture was 55 minutes long and divided into 10-minute ``stages'' (based on the average human attention span \cite{mckeachie2006teaching}). A 10-minute 2D video established baseline EEG signals before the experiment. Participants completed quizzes (10 multiple-choice questions) after each module to assess learning performance.

\item \textbf{System Design:} The EEG data was analyzed in 10-minute intervals to assess alertness and readiness to learn~\cite{taherisadr2023erudite}. The system structure is similar to the thermal house model (Application 1), with the human state inferred at the edge before an action is sent to the cloud.

\end{itemize}

\vspace{-2mm}
\subsection{Human-Centric Design and MDP Formulation}

\begin{figure*}[!t]
\begin{minipage}{0.4\textwidth}
    \centering
    \includegraphics[trim={0 0.5cm 0 0},clip,width=0.9\textwidth]{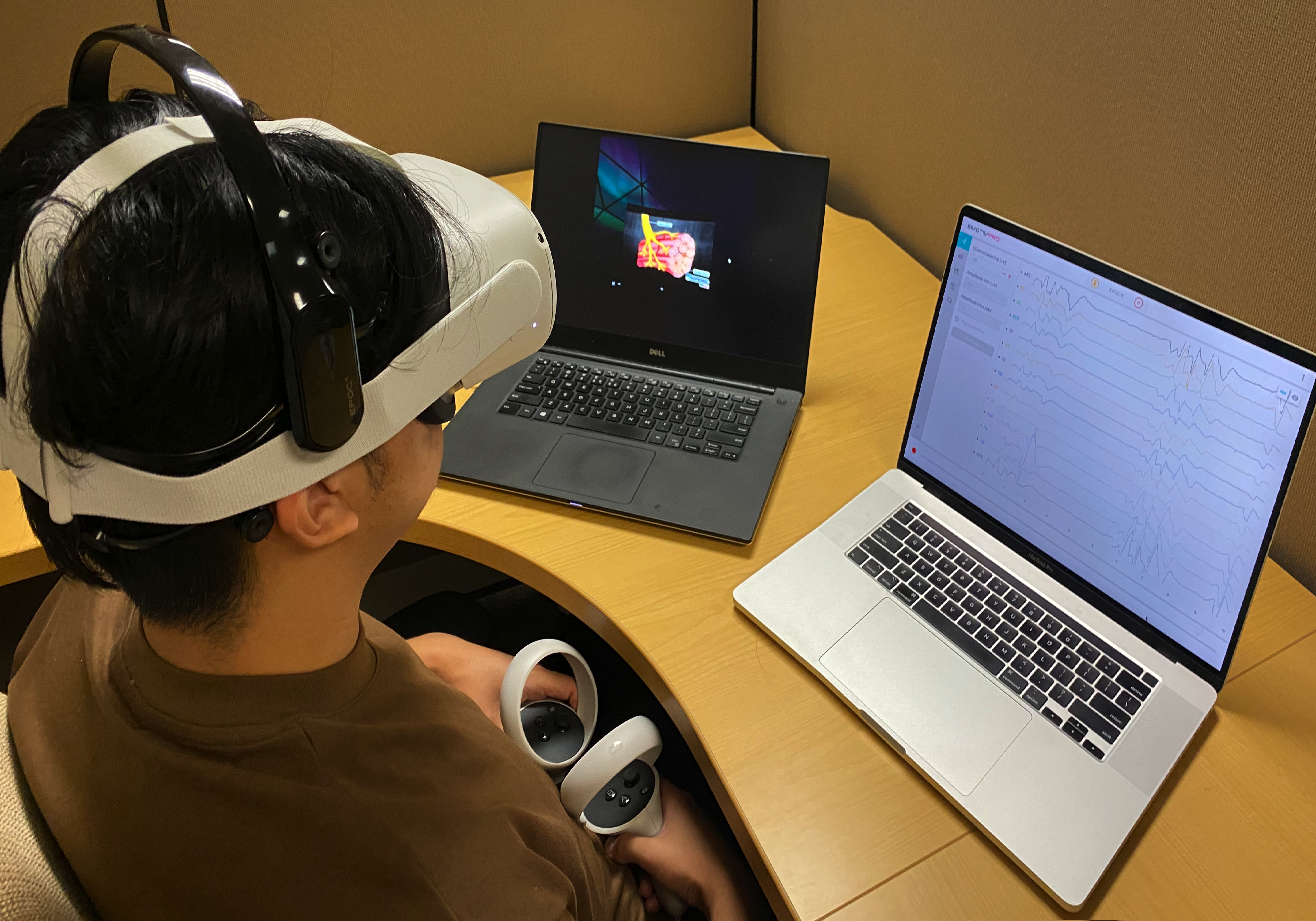}\vspace{-3mm}
    \caption{EMOTIV and Oculus on a participant.}
    \label{fig:emotiv}
\end{minipage}\hfill
\begin{minipage}{0.6\textwidth}
\centering
\captionof{table}{Human state is one of 8 states depending on the alertness level (AL), fatigue level (FL), and vertigo level (VL).}
    \label{tab:state}
\resizebox{0.8\textwidth}{!}{
    \begin{tabular}{
    >{\columncolor[HTML]{cbcfd1}}c 
    >{\columncolor[HTML]{cbcfd1}}c 
    >{\columncolor[HTML]{cbcfd1}}c 
    >{\columncolor[HTML]{cbcfd1}}c 
    >{\columncolor[HTML]{cbcfd1}}c 
    >{\columncolor[HTML]{cbcfd1}}c 
    >{\columncolor[HTML]{cbcfd1}}c 
    >{\columncolor[HTML]{cbcfd1}}c 
    >{\columncolor[HTML]{cbcfd1}}c}
    \textbf{AL}  & \cellcolor[HTML]{00FF00}1 & \cellcolor[HTML]{00FF00}1  & \cellcolor[HTML]{00FF00}1  & \cellcolor[HTML]{00FF00}1   & \cellcolor[HTML]{ADD8E6}0 & \cellcolor[HTML]{ADD8E6}0 & \cellcolor[HTML]{ADD8E6}0  & \cellcolor[HTML]{ADD8E6}0  \\
    
    \textbf{FL} & \cellcolor[HTML]{00FF00}1 & \cellcolor[HTML]{00FF00}1 & \cellcolor[HTML]{ADD8E6}0 & \cellcolor[HTML]{ADD8E6}0 & \cellcolor[HTML]{00FF00}1 & \cellcolor[HTML]{00FF00}1 & \cellcolor[HTML]{ADD8E6}0 & \cellcolor[HTML]{ADD8E6}0  \\
    
    \textbf{VL} & \cellcolor[HTML]{00FF00}1 & \cellcolor[HTML]{ADD8E6}0 & \cellcolor[HTML]{00FF00}1 & \cellcolor[HTML]{ADD8E6}0 & \cellcolor[HTML]{00FF00}1 & \cellcolor[HTML]{ADD8E6}0 & \cellcolor[HTML]{00FF00}1 & \cellcolor[HTML]{ADD8E6}0  \\
    
    \textbf{State} & \cellcolor[HTML]{cbcfd1}\textbf{S8} & \cellcolor[HTML]{cbcfd1}\textbf{S7} & \cellcolor[HTML]{cbcfd1}\textbf{S6} & \cellcolor[HTML]{cbcfd1}\textbf{S5} & \cellcolor[HTML]{cbcfd1}\textbf{S4} & \cellcolor[HTML]{cbcfd1}\textbf{S3} & \cellcolor[HTML]{cbcfd1}\textbf{S2} & \cellcolor[HTML]{cbcfd1}\textbf{S1}
    \end{tabular}
    } 
    \vspace{0.3cm}
    
\end{minipage}\hfill
\end{figure*}

The application monitors fluctuations in the student's mental and physiological state, which are critical for learning outcomes. The system is designed as an edge-based CPS application where sensitive data is processed locally, and only the control actions are exposed to the cloud. 
The control task is modeled as an MDP: 
\begin{itemize}[noitemsep, topsep=0pt, leftmargin=*]

\item \textbf{Private State ($s_\text{private}$)}: This is the sensitive data measured by EEG (Figure \ref{fig:emotiv}) and categorized into a discrete space $\mathbf{S}$. We define the state based on three binary features: Alertness Level (AL), Fatigue Level (FL), and Vertigo Level (VL). This results in eight discrete states ($S_1$ to $S_8$) (Table \ref{tab:state}), where $S_8$ is the optimal state (Alert, No Fatigue, No Vertigo)~\cite{taherisadr2023eruditee}.

\item \textbf{Action ($\mathcal{A}^*$)}: The exposed action space comprises five discrete control signals sent to the cloud, such as (1) provide a break, (2) enable VR mode (2D to 3D), (3) disable VR mode (3D to 2D), (4) alter content, or (5) maintain the current state.

\item \textbf{Reward ($\mathcal{R}_i$)}: The reward is primarily governed by the utility metric: the individual's performance on quizzes (expressed as a percentage) taken after each learning module.
\end{itemize}

\begin{figure*}[!t]
\begin{minipage}{0.8\textwidth}
    \centering
    \includegraphics[scale=0.4]{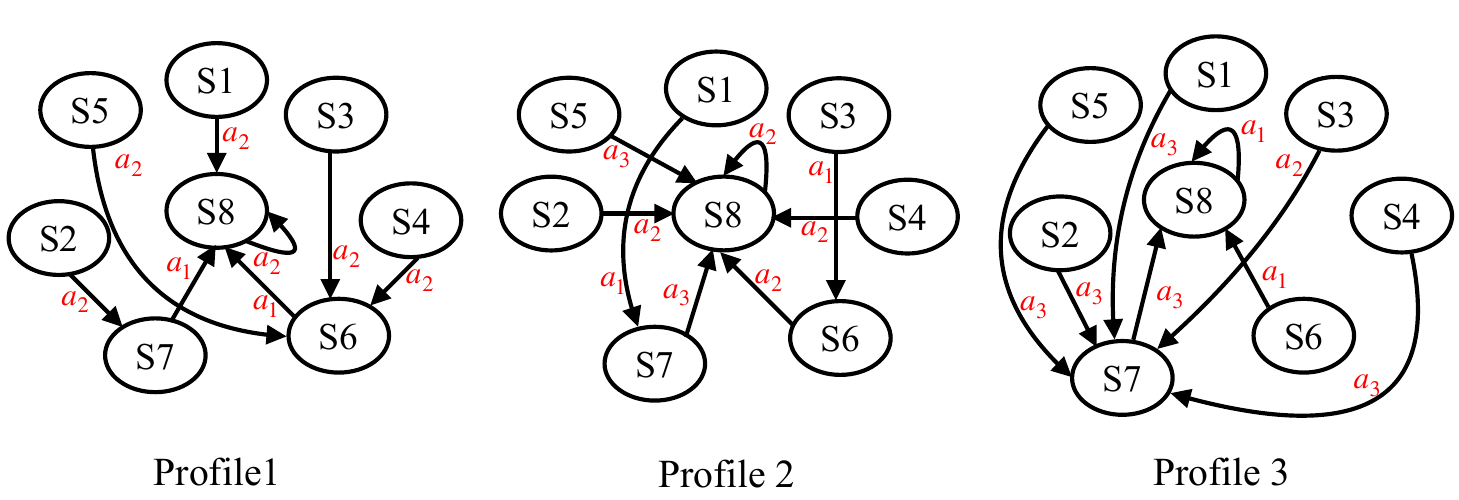}\vspace{-3mm}
    \caption{MDP for three human profiles in VR learning environment.}
    \label{fig:stateSpace}
\end{minipage}
\end{figure*}

The 15 participants in the VR experiment were classified into three distinct groups, or profiles ($P_1,P_2,P_3$), based on their behavioral characteristics and tolerance to VR exposure, reflecting the variability inherent in human-centric systems. The behavior of each profile was subsequently modeled using a separate Markov Decision Process (MDP) (Figure \ref{fig:stateSpace}):
\begin{itemize}[noitemsep, leftmargin=*, topsep=0pt]
\item \textbf{Profile 1 ($P_1$ - High Tolerance)}: This group consists of individuals who exhibit high VR tolerance. They do not experience simulator sickness or vertigo symptoms even after extended VR use (exceeding 20 minutes). Their policies prioritize maintaining the engaging VR mode for longer periods.
\item \textbf{Profile 2 ($P_2$ - Moderate Tolerance)}: This includes individuals who may experience cybersickness or fatigue during VR exposure. Their behavior is characterized by a moderate likelihood of transitioning to less favorable states, requiring the RL agent to intermittently intervene (switch to 2D mode or provide a break).

\item \textbf{Profile 3 ($P_3$ - Low Tolerance)}: This group comprises individuals with the lowest VR tolerance. They are highly susceptible to vertigo and fatigue, necessitating frequent, proactive intervention by the RL agent to switch presentation modes or provide breaks to maintain the learning state.
\end{itemize}
	
The RL agent's objective is to maximize the utility (quiz score) for each personalized profile by learning the optimal action sequence, while \sysname ensures that this personalized control does not excessively compromise the privacy of the underlying  state.

\vspace{-2mm}
\subsection{\sysname Adaptation and Evaluation}

We follow the evaluation approach as the first simulated HVAC application. We used a 10-layer EE-DQN architecture, with separate three networks trained for each of the three behavioral profiles.

\vspace{-2mm}
\subsubsection{Architectural Baseline}
The PHASE 1 (Sequential Training) established the initial Q-values. Analysis of Profile 1 ($P_1$) showed that the Mutual Information (MI) peaked at exit branch $L_4$ (Figure~\ref{fig:miapp2}). This peak MI confirmed that the DQN depth required for optimal learning performance (utility) creates the maximum information leakage about the private human state.
\vspace{-1mm}
\subsubsection{\textbf{Privacy Leak Assessment}}
Assuming an ``honest-but-curious'' cloud, we quantified the baseline information leakage by applying unsupervised clustering to the time-series of exposed actions ($\mathcal{A}^*$). The cloud achieved 82\% accuracy in inferring the participants' private mental state by clustering the control signals which is the baseline inference accuracy. It is crucial to note that even if the cloud lacks direct knowledge of the exact human state, changes in the human state (or learning patterns) due to observed actions can be inferred. This inference could potentially reveal sensitive information, such as the participant's attention span, given that the cloud has prior knowledge of the application domain. This validates the need for \sysname's mitigation strategies.
\vspace{-1mm}
\subsubsection{\textbf{\sysname Personalized Adaptation }}
We initiated PHASE 2 to train the $UCL$ and $PCL$ paths to enforce the personalized budgets $u$ and $p$ (Algorithm~\ref{alg:algorithms}).
\begin{itemize}[noitemsep, topsep=0pt, leftmargin=*]
\item Budget Constrained Exit Selection (Algorithm~\ref{alg:exit_branches_inference}): Table \ref{table:allexitalow2} shows the layers where the exit branch $E_{k^*}$ satisfies both budgets ($UCL=1$ and $PCL=1$). This table confirms that the permissible exit depths are personalized to each profile's VR tolerance and shows that stricter budgets (high $u$, high $p$) often result in no compliant exits ($x$).

\item Mitigation Results: By employing Algorithm~\ref{alg:exit_branches_inference} to select the earliest exit compliant with $u=0.75$ and $p=0.7$, the state prediction accuracy achieved by the cloud was significantly reduced. On average, across all profiles, privacy (reduction in state prediction accuracy) improved by 28\%, corresponding to a controlled utility (quiz performance) drop of 28\%.
\end{itemize}

\vspace{-2mm}
\subsubsection{\textbf{Privacy-Utility Tradeoff}}
\begin{figure*}
\begin{minipage}{0.45\textwidth}
    \centering
    \includegraphics[width=0.9\linewidth]{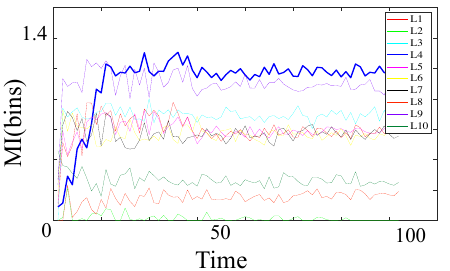} 
    \caption{MI for all exit branches of profile~1.}
    \label{fig:miapp2}
\end{minipage}\hfill
\begin{minipage}{0.45\textwidth}
\centering
\includestandalone[width=\linewidth]{figures/app2/app2put}\\
\caption{Privacy-utility tradeoff for three profiles with various values of $\mathbf{u}$ at $\mathbf{p=0.7}$.}
\label{fig:put2}
\end{minipage}
\end{figure*}

The analysis in Figure~\ref{fig:put2} confirms the clear negative correlation between the two objectives: increasing the utility budget $u$ compromises privacy (higher clustering accuracy). The system provides a mechanism for the user to select their preferred operating point on this tradeoff curve.

In summary, this deployment in VR context with real human data confirms that \sysname successfully achieves personalized and dynamic control over the privacy-utility tradeoff in a complex, human-centric learning environment by enforcing $UCL$ and $PCL$ constraints, which demonstrates its viability for real-world adaptive CPHS systems.

\section{Summary of Results}

Across both applications, the average privacy enhancement is 31\% derived from the 26\%, 23\%, and 21\% improvements for H1, H2, H3 in HVAC and the average 28\% improvement across VR profiles, collectively averaging to approximately 25.67\%. 

The sensitivity of exit selection to budget parameters is directly visible in Tables~\ref{table:allexitalow} and~\ref{table:allexitalow2}: small changes in $u$ or $p$ can shift the set of compliant exits significantly, particularly near the boundary conditions where performance or privacy is tightly constrained. Practitioners are encouraged to use these tables as pre-deployment configuration guides. 

The budgets $u$ and $p$ are user-defined parameters that can be elicited through several practical mechanisms: (i) explicit preference questionnaires where users specify acceptable performance degradation and privacy risk levels; (ii) application defaults based on regulatory requirements (e.g., HIPAA-compliant settings for healthcare CPS); or (iii) adaptive elicitation via short trials where the user rates the tradeoff under different budget presets. Since \sysname's Phase 2 training is budget-agnostic (any ($u$,$p$) combinations are encoded via the UCL/PCL label scheme), the system can support runtime budget changes without retraining by re-evaluating Algorithm~\ref{alg:algorithms} with the new budgets on the already-trained confidence paths. Future work will explore user-centered interfaces for dynamic budget elicitation and robust budget estimation that accounts for uncertainty in the user's stated preferences.




\begin{figure*}[!t]
\begin{minipage}{0.95\textwidth}
\captionof{table}{$\mathbf{PCL}$ and $\mathbf{UCL}$ confidence per layer for all profiles at various values of utility budget $\mathbf{u}$ and privacy budget $\mathbf{p}$.}
\label{table:allexitalow2}
\resizebox{0.95\textwidth}{!}{
\begin{tabular}{|c||c|c|c||c|c|c||c|c|c||c|c|c||}
        \hline
         &  \multicolumn{3}{c||}{\cellcolor[HTML]{ADD8E6}u = 0.85} & \multicolumn{3}{c||}{\cellcolor[HTML]{ADD8E6}u = 0.75}   & \multicolumn{3}{c||}{\cellcolor[HTML]{ADD8E6}u = 0.65}  & \multicolumn{3}{c||}{\cellcolor[HTML]{ADD8E6}u = 0.55}  \\
        \cline{1-13}
         Human  & $P_1$ & $P_2$ & $P_3$  &  $P_1$ & $P_2$ & $P_3$  &  $P_1$ & $P_2$ & $P_3$  &  $P_1$ & $P_2$ & $P_3$  \\
        \hline
        \cellcolor[HTML]{ADD8E6}p = 0.9   & $\times$  & $\times$  & $\times$  & $\times$  & $\times$  & $\times$  & $\times$  & $\times$  & $\times$  & $\times$   & $L_{5}$  &  $L_{7,8}$   \\
        \hline
        
        \cellcolor[HTML]{ADD8E6}p = 0.8   & $\times$  & $\times$  & $\times$ 
        & $L_4$   &  $L_5$  & $L_7$ & 
        $L_{2,4}$  & $L_{5,7}$   &  $L_{6,7}$  
        & $L_{2,4,8}$  & $L_{1,5,8,}$  &  $L_{2,6,7}$    \\
        \hline
        
        \cellcolor[HTML]{ADD8E6}p = 0.7   & $\times$  & $\times$  & $\times$  & 
        $L_{3,4,6}$  & $L_{5,9}$  &  $L_{3,5,6}$ & 
        $L_{4,7}$  & $L_{4,5,9}$   &  $L_{4,7}$  & 
        $L_{1,6}$  & $L_{3,6,8}$  &  $L_{3,5,7}$    \\
        \hline
        \cellcolor[HTML]{ADD8E6}p = 0.6   & $\times$   & $\times$  & $L_{3}$  & $L_{1,6,7}$  & $L_{4,5}$  &  $L_{4,5,9}$ & 
        $L_{2,4,10}$  & $L_{5,8,9}$   &  $L_{4,7}$  & 
        $L_{1,4,9}$  & $L_{4,5,7}$  &  $L_{7,8,10}$\\
        \hline
    \end{tabular}
}

\end{minipage}
\end{figure*}

%% file: figures/app2/app2put.tex
\begin{tikzpicture}[font=\large\bfseries]
\begin{axis}[
    xlabel={Utility budget},
    ylabel={Performance Drop (\%)},
    xmin=0.63, xmax=0.87,
    ymin=0.15, ymax=0.95,
    xtick={0.65, 0.75, 0.85},
    ytick={0.2, 0.3, 0.4, 0.5, 0.6, 0.7, 0.8, 0.9, 1.0},
    legend style={at={(1.2,1)}, anchor= north west},
    y axis line style={-},
    x axis line style={-},
    ylabel near ticks,
    grid=major
    ]

    \addplot[mark=*, blue, dashed] coordinates {(0.65, 0.58) (0.75, 0.5) (0.85, 0.2)}; 
    \addplot[mark=*, red, dashed] coordinates {(0.65, 0.6) (0.75, 0.3) (0.85, 0.28)}; 
    \addplot[mark=*, Green, dashed] coordinates {(0.65, 0.42) (0.75, 0.32) (0.85, 0.25)}; 
    
    \legend{Perf. Drop (P1), Perf. Drop (P2), Perf. Drop (P3)}
\end{axis}

\begin{axis}[
    axis y line*=right,
    axis x line=none,
    ylabel={Privacy Leak (\%)},
    xmin=0.63, xmax=0.87,
    ymin=0.15, ymax=0.95,
    ytick={0.2, 0.3, 0.4, 0.5, 0.6, 0.7, 0.8, 0.9, 1.0},
    legend style={at={(1.2,0.7)}, anchor= north west},
    y axis line style={-},
    x axis line style={-},
    ylabel near ticks,
    grid=major
    ]
    \addplot[mark=x, blue] coordinates {(0.65, 0.63) (0.75, 0.64) (0.85, 0.83)}; 
    \addplot[mark=x, red] coordinates {(0.65, 0.63) (0.75, 0.75) (0.85, 0.79)}; 
    \addplot[mark=x, Green] coordinates {(0.65, 0.64) (0.75, 0.65) (0.85, 0.78)}; 
    
    \legend{Privacy Leak (P1), Privacy Leak (P2), Privacy Leak (P3)}
\end{axis}

\end{tikzpicture}

%% file: 08_discussion.tex
\vspace{-2mm}
\section{Discussion and Limitations}\label{sec:discuss}

This section addresses the novelty and design rationale of \sysname while  discussing the inherent limitations of privacy quantification and the socio-technical dimensions of the privacy-utility tradeoff.
\vspace{-2mm}
\paragraph{\textbf{Novelty and Design Rationale}}
The \sysname framework represents a significant advance in personalized privacy for human-centric Cyber-Physical Systems (CPHS) by integrating Early-Exit Deep Reinforcement Learning (EE-DRL) with a novel privacy mechanism. This achieves simultaneous personalization, dynamic privacy enforcement, and utility maximization—a capability absent in prior work. The core innovation is the Privacy Confidence Label (PCL) path, explicitly trained using Mutual Information (MI), which enables the DRL agent to dynamically manage privacy by leveraging the inherent structural redundancy of the EE network. DRL is essential because the optimal policy $\pi(s,a)$ is non-stationary and personalized; given the requirement for continuous adaptation (driven by Algorithm~\ref{alg:inference_with_retraining}) and a personalized objective function (set by $u$ and $p$ budgets), static optimization methods are computationally infeasible for continuous human-centric adaptation.

We acknowledge that the relationship between exit depth and MI is established empirically in this work. While the data processing inequality implies that deeper representations cannot contain less information than shallower ones in general, a formal proof that \sysname's specific sequential training procedure monotonically reduces MI with decreasing exit index is beyond the scope of this paper and represents important future theoretical work. 


\vspace{-2mm}
\paragraph{\textbf{Privacy Quantification}} We recognize the inherent challenges in quantifying information leakage, but assert the practical validity of our mitigation strategy. We utilize Mutual Information (MI) as a principled, application-agnostic metric for leakage. While real-world MI estimation is subject to finite data and computational constraints, our empirical results confirm that the MI-based PCL is an effective and robust mechanism for practical privacy mitigation under the ``honest-but-curious'' threat model. The framework successfully reduced the accuracy of inference (clustering) by an average of $31\%$ across two distinct applications. Though we acknowledge that MI may not capture all socio-technical nuances of privacy, this high degree of quantifiable reduction validates the PCL mechanism's ability to enforce personalized privacy bounds.

Furthermore, MI is estimated per step over a window rather than over full trajectories $(I(A_{1:T}; S_{1:T}))$, which is a known simplification. For our discrete, finite applications this approximation is reasonable, but extending \sysname to continuous high-dimensional settings would require more sophisticated sequential MI estimators.

\paragraph{\textbf{Privacy vs. Utility Tradeoff}} 
A known limitation of the current fallback policy is that privacy may be temporarily violated under stringent utility demands. Hard privacy constraints (never violating PCL) could theoretically be enforced by rejecting actions and substituting a privacy-safe default action, but this could compromise physical safety (e.g., HVAC setpoint defaults in extreme temperatures). We treat this as an application-level tradeoff and note that our evaluation results ~\ref{table:allexitalow} and ~\ref{table:allexitalow2} can serve as pre-deployment tools: system designers can select budget pairs that avoid 'x' entries, ensuring zero fallback frequency for the chosen operating point.

\paragraph{\textbf{Edge Deployment}} We acknowledge that formal profiling on representative edge hardware (e.g., Raspberry Pi, NVIDIA Jetson) would strengthen the deployment argument. However, the long action periods in our applications (1 hour for HVAC, 10 minutes for VR) mean that inference latency is not a binding constraint. Retraining (Algorithm~\ref{alg:inference_with_retraining}) is triggered only upon significant behavioral drift and runs offline over extended periods (approximately 10 days in the HVAC simulation), making it compatible with background computation on edge devices. We treat systematic latency benchmarking as future work for tighter real-time CPS applications.

\vspace{-2mm}
\paragraph{\textbf{Perception of Privacy}}  Our findings confirm the necessary privacy-utility tradeoff: enhancing privacy comes at a measurable cost to utility. This tradeoff is characterized by an average utility loss of 10–16\% in exchange for the privacy protection. Crucially, this cost is both acceptable and controllable for privacy-conscious users. 
Looking ahead, we plan to address the socio-technical dimensions of privacy by exploring how different contextual and behavioral definitions of privacy influence the development of new safeguards\cite{ali2020image}. 

%% file: 09_conclusion.tex
\vspace{-3mm}
\section{Conclusion}\label{sec:conclusion}


This paper introduced \sysname, a novel framework that uses an Early-Exit Deep Reinforcement Learning (EE-DRL) architecture to dynamically manage the personalized privacy-utility tradeoff in human-centric Cyber-Physical Systems. Our core innovation lies in the Privacy Confidence Label (PCL) path, which is trained using Mutual Information (MI) as a quantifiable penalty to enable fine-grained, dynamic privacy decisions at the edge. We validated \sysname's efficacy across two distinct, real-world relevant applications: a personalized HVAC smart home system and a smart VR classroom environment, successfully adapting the control policy to individual behavioral profiles (e.g., thermal comfort and VR sickness tolerance). The results confirm that \sysname provides dynamic control that strictly enforces user-defined utility ($u$) and privacy ($p$) budgets, demonstrating a significant defense against adversarial inference. By selecting the earliest compliant exit layer in the DRL agent, \sysname successfully improved the resilience against state prediction by an average of 25.67\% across both applications. Ultimately, \sysname offers a robust and viable paradigm for engineering future intelligent CPS that prioritize user privacy without sacrificing the necessary policy adaptability required for optimizing human-centric services.

%% file: references.bib
@inproceedings{niu2021adapdp,
  title={AdaPDP: Adaptive personalized differential privacy},
  author={Niu, Ben and Chen, Yahong and Wang, Boyang and Wang, Zhibo and Li, Fenghua and Cao, Jin},
  booktitle={IEEE INFOCOM 2021-IEEE conference on computer communications},
  pages={1--10},
  year={2021},
  organization={IEEE}
}

@article{fallah2020personalized,
  title={Personalized federated learning with theoretical guarantees: A model-agnostic meta-learning approach},
  author={Fallah, Alireza and Mokhtari, Aryan and Ozdaglar, Asuman},
  journal={Advances in neural information processing systems},
  volume={33},
  pages={3557--3568},
  year={2020}
}

@inproceedings{xiao2023micpro,
  title={MicPro: Microphone-based Voice Privacy Protection},
  author={Xiao, Shilin and Ji, Xiaoyu and Yan, Chen and Zheng, Zhicong and Xu, Wenyuan},
  booktitle={Proceedings of the 2023 ACM SIGSAC Conference on Computer and Communications Security},
  pages={1302--1316},
  year={2023}
}

@inproceedings{patel2022model,
  title={Model explanations with differential privacy},
  author={Patel, Neel and Shokri, Reza and Zick, Yair},
  booktitle={Proceedings of the 2022 ACM Conference on Fairness, Accountability, and Transparency},
  pages={1895--1904},
  year={2022}
}

@inproceedings{ito2022success,
  title={On the success rate of side-channel attacks on masked implementations: information-theoretical bounds and their practical usage},
  author={Ito, Akira and Ueno, Rei and Homma, Naofumi},
  booktitle={Proceedings of the 2022 ACM SIGSAC Conference on Computer and Communications Security},
  pages={1521--1535},
  year={2022}
}

@article{Zhao2026FinP,
  author  = {Zhao, Tianyu and Srewa, Mahmoud and Elmalaki, Salma},
  title   = {FinP: Fairness-in-Privacy in Federated Learning by Addressing Disparities in Privacy Risk},
  journal = {Proceedings on Privacy Enhancing Technologies},
  year    = {2026},
  volume  = {2026},
  number  = {4},
  pages   = {587--607},
  doi     = {10.56553/popets-2026-0136}
}

@inproceedings{li2022auditing,
  title={Auditing membership leakages of multi-exit networks},
  author={Li, Zheng and Liu, Yiyong and He, Xinlei and Yu, Ning and Backes, Michael and Zhang, Yang},
  booktitle={Proceedings of the 2022 ACM SIGSAC Conference on Computer and Communications Security},
  pages={1917--1931},
  year={2022}
}

@inproceedings{teerapittayanon2016branchynet,
  title={Branchynet: Fast inference via early exiting from deep neural networks},
  author={Teerapittayanon, Surat and McDanel, Bradley and Kung, Hsiang-Tsung},
  booktitle={2016 23rd international conference on pattern recognition (ICPR)},
  pages={2464--2469},
  year={2016},
  organization={IEEE}
}

@article{matsubara2022split,
  title={Split computing and early exiting for deep learning applications: Survey and research challenges},
  author={Matsubara, Yoshitomo and Levorato, Marco and Restuccia, Francesco},
  journal={ACM Computing Surveys},
  volume={55},
  number={5},
  pages={1--30},
  year={2022},
  publisher={ACM New York, NY}
}

@incollection{hlavka2020security,
  title={Security, privacy, and information-sharing aspects of healthcare artificial intelligence},
  author={Hl{\'a}vka, Jakub P},
  booktitle={Artificial Intelligence in Healthcare},
  pages={235--270},
  year={2020},
  publisher={Elsevier}
}

@article{omonkhoa2021review,
  title={A review of the issues and challenges in IoT security using machine learning techniques},
  author={Omonkhoa, Liberty Adams},
  year={2021},
  publisher={Alt{\i}nba{\c{s}} {\"U}niversitesi/Lisans{\"u}st{\"u} E{\u{g}}itim Enstit{\"u}s{\"u}}
}

@techreport{IEEE_CSS_Roadmap_2023,
  author    = {Annaswamy, A. M. and Johansson, K. H. and Pappas, G. J.},
  editor    = {Annaswamy, A. M. and Johansson, K. H. and Pappas, G. J.},
  title     = {{Control for Societal-scale Challenges: Road Map 2030}},
  institution = {IEEE Control Systems Society Publication},
  year      = {2023},
  type      = {Road Map},
  note      = {The editors are also listed as authors for indexing purposes.},
  url       = {https://ieeecss.org/sites/ieeecss/files/2023-05/CSS_Roadmap_2023_Cover_8x10_FINAL_REV5.pdf},
  month     = {May}
}

@article{taherisadr2023erudite,
  title={Erudite: Human-in-the-loop iot for an adaptive personalized learning system},
  author={Taherisadr, Mojtaba and Faruque, Mohammad Abdullah Al and Elmalaki, Salma},
  journal={arXiv preprint arXiv:2303.04292},
  year={2023}
}

@article{taherisadr2023eruditee,
  title={Erudite: Human-in-the-loop iot for an adaptive personalized learning system},
  author={Taherisadr, Mojtaba and Al Faruque, Mohammad Abdullah and Elmalaki, Salma},
  journal={IEEE Internet of Things Journal},
  year={2023},
  publisher={IEEE}
}

@inproceedings{kosta2022rapid,
  title={RAPID-RL: A Reconfigurable Architecture with Preemptive-Exits for Efficient Deep-Reinforcement Learning},
  author={Kosta, Adarsh Kumar and Anwar, Malik Aqeel and Panda, Priyadarshini and Raychowdhury, Arijit and Roy, Kaushik},
  booktitle={2022 International Conference on Robotics and Automation (ICRA)},
  pages={7492--7498},
  year={2022},
  organization={IEEE}
}

@article{taherisadr2023adaparl,
  title={adaparl: Adaptive privacy-aware reinforcement learning for sequential-decision making human-in-the-loop systems},
  author={Taherisadr, Mojtaba and Stavroulakis, Stelios Andrew and Elmalaki, Salma},
  journal={arXiv preprint arXiv:2303.04257},
  year={2023}
}

@inproceedings{elmalaki2021fair,
  title={Fair-iot: Fairness-aware human-in-the-loop reinforcement learning for harnessing human variability in personalized iot},
  author={Elmalaki, Salma},
  booktitle={Proceedings of the International Conference on Internet-of-Things Design and Implementation},
  pages={119--132},
  year={2021}
}

@book{sadigh2017active,
  title={Active preference-based learning of reward functions},
  author={Sadigh, Dorsa and Dragan, Anca D and Sastry, Shankar and Seshia, Sanjit A},
  year={2017}
}

@inproceedings{elmalaki2022maconauto,
  title={MAConAuto: Framework for Mobile-Assisted Human-in-the-Loop Automotive System},
  author={Elmalaki, Salma},
  booktitle={2022 IEEE Intelligent Vehicles Symposium (IV)},
  pages={740--749},
  year={2022},
  organization={IEEE}
}

@article{hadfield2016cooperative,
  title={Cooperative inverse reinforcement learning},
  author={Hadfield-Menell, Dylan and Russell, Stuart J and Abbeel, Pieter and Dragan, Anca},
  journal={Advances in neural information processing systems},
  volume={29},
  year={2016}
}

@inproceedings{elmalaki2018sentio,
  title={Sentio: Driver-in-the-loop forward collision warning using multisample reinforcement learning},
  author={Elmalaki, Salma and Tsai, Huey-Ru and Srivastava, Mani},
  booktitle={Proceedings of the 16th ACM Conference on Embedded Networked Sensor Systems},
  pages={28--40},
  year={2018}
}

@inproceedings{ahadi2021adas,
  title={Adas-rl: Adaptive vector scaling reinforcement learning for human-in-the-loop lane departure warning},
  author={Ahadi-Sarkani, Armand and Elmalaki, Salma},
  booktitle={Proceedings of the First International Workshop on Cyber-Physical-Human System Design and Implementation},
  pages={13--18},
  year={2021}
}

@article{eslinger1985severe,
  title={Severe disturbance of higher cognition after bilateral frontal lobe ablation: patient EVR},
  author={Eslinger, Paul J and Damasio, Antonio R},
  journal={Neurology},
  volume={35},
  number={12},
  pages={1731--1731},
  year={1985},
  publisher={AAN Enterprises}
}

@inproceedings{wu2023enhancing,
  title={Enhancing Privacy in Federated Learning via Early Exit},
  author={Wu, Yashuo and Chiasserini, Carla Fabiana and Malandrino, Francesco and Levorato, Marco},
  booktitle={Proceedings of the 5th workshop on Advanced tools, programming languages, and PLatforms for Implementing and Evaluating algorithms for Distributed systems},
  pages={1--5},
  year={2023}
}

@misc{nestcali,
author = {Greentech Media},
title = {Nest Home},
howpublished = {\url{https://vaeec.org/news/inside-nests-/50000-home-virtual-power-plant-for-southern-california-edison/}},
note = {Accessed: 2024-09-26},
year = {2022}
}

@misc{MATLABther,
  author =       "MATLAB",
  year =         "2022",
  month =        mar,
  title =        "Thermal Model of a House",
  lastaccessed = "June 10, 2024",
  url =          "https://www.mathworks.com/help/simulink/slref/thermal-model-of-a-house.html",
}

@article{ali2020image,
  title={Image versus information: Changing societal norms and optimal privacy},
  author={Ali, S Nageeb and B{\'e}nabou, Roland},
  journal={American Economic Journal: Microeconomics},
  volume={12},
  number={3},
  pages={116--164},
  year={2020},
  publisher={American Economic Association 2014 Broadway, Suite 305, Nashville, TN 37203-2425}
}

@inproceedings{zhang2023facereader,
  title={FaceReader: Unobtrusively Mining Vital Signs and Vital Sign Embedded Sensitive Info via AR/VR Motion Sensors},
  author={Zhang, Tianfang and Ye, Zhengkun and Mahdad, Ahmed Tanvir and Akanda, Md Mojibur Rahman Redoy and Shi, Cong and Wang, Yan and Saxena, Nitesh and Chen, Yingying},
  booktitle={Proceedings of the 2023 ACM SIGSAC Conference on Computer and Communications Security},
  pages={446--459},
  year={2023}
}

@misc{rushh,
  author =       "Edison",
  year =         "2024",
  month =        mar,
  title =        "Smart Energy Program",
  lastaccessed = "Sept 25, 2024",
  url =          "https://www.sce.com/tnc/smart-energy-program-terms-and-conditions",
}

@misc{Globe,
  author =       "campbellsci",
  year =         "2024",
  month =        mar,
  title =        "black globe thermometer",
  lastaccessed = "June 10, 2024",
  url =          "https://www.campbellsci.com/blackglobe",
}

@misc{KhanBiology,
author = {KhanAcademy},
title = {AP College Biology},
howpublished = {\url{https://www.khanacademy.org/science/ap-biology}},
note = {Accessed: 2024-10-07},
year = {2024}
}

@misc{KhanPhysics,
author = {KhanAcademy},
title = {AP College Physics},
howpublished = {\url{https://www.khanacademy.org/science/ap-physics-2}},
note = {Accessed: 2024-10-07},
year = {2024}
}

@inproceedings{terai2020detecting,
  title={Detecting Learner Drowsiness Based on Facial Expressions and Head Movements in Online Courses},
  author={Terai, Shogo and Shirai, Shizuka and Alizadeh, Mehrasa and Kawamura, Ryosuke and Takemura, Noriko and Uranishi, Yuki and Takemura, Haruo and Nagahara, Hajime},
  booktitle={Proceedings of the 25th International Conference on Intelligent User Interfaces Companion},
  pages={124--125},
  year={2020}
}

@misc{KhanChemistry,
author = {KhanAcademy},
title = {Organic Chemistry},
howpublished = {\url{https://www.khanacademy.org/science/organic-chemistry}},
note = {Accessed: 2024-10-07},
year = {2024}
}

@article{fraenkel2004complexity,
  title={Complexity, appeal and challenges of combinatorial games},
  author={Fraenkel, Aviezri S},
  journal={Theoretical Computer Science},
  volume={313},
  number={3},
  pages={393--415},
  year={2004},
  publisher={Elsevier}
}

@article{gerber2014energyplus,
  title={energyplus energy Simulation Software},
  author={Gerber, Michael},
  year={2014}
}

@article{zhang2022homomorphic,
  title={Homomorphic Encryption-based Privacy-preserving Federated Learning in IoT-enabled Healthcare System},
  author={Zhang, Li and Xu, Jianbo and Vijayakumar, Pandi and Sharma, Pradip Kumar and Ghosh, Uttam},
  journal={IEEE Transactions on Network Science and Engineering},
  year={2022},
  publisher={IEEE}
}

@article{shany2012sensors,
  title={Sensors-based wearable systems for monitoring of human movement and falls},
  author={Shany, Tal and Redmond, Stephen J and Narayanan, Michael R and Lovell, Nigel H},
  journal={IEEE Sensors Journal},
  volume={12},
  number={3},
  pages={658--670},
  year={2012},
  publisher={IEEE}
}

@article{wang2019privacy,
  title={Privacy-preserving q-learning with functional noise in continuous spaces},
  author={Wang, Baoxiang and Hegde, Nidhi},
  journal={Advances in Neural Information Processing Systems},
  volume={32},
  year={2019}
}

@inproceedings{jung2017towards,
  title={Towards integration of doppler radar sensors into personalized thermoregulation-based control of HVAC},
  author={Jung, Wooyoung and Jazizadeh, Farrokh},
  booktitle={Proceedings of the 4th ACM International Conference on Systems for Energy-Efficient Built Environments},
  pages={21},
  year={2017},
  organization={ACM}
}

@article{barrett1993sleep,
  title={The sleep-evoked decrease of body temperature},
  author={Barrett, Judith and Lack, Leon and Morris, Mary},
  journal={Sleep},
  volume={16},
  number={2},
  pages={93--99},
  year={1993},
  publisher={Oxford University Press}
}

@article{lim2008human,
  title={Human thermoregulation and measurement of body temperature in exercise and clinical settings},
  author={Lim, Chin Leong and Byrne, Chris and Lee, Jason KW},
  journal={Annals Academy of Medicine Singapore},
  volume={37},
  number={4},
  pages={347},
  year={2008}
}

@book{carroll2006elsevier,
  title={Elsevier's Integrated Physiology E-Book},
  author={Carroll, Robert G},
  year={2006},
  publisher={Elsevier Health Sciences}
}

@inproceedings{nguyen2016lightweight,
  title={A lightweight and inexpensive in-ear sensing system for automatic whole-night sleep stage monitoring},
  author={Nguyen, Anh and Alqurashi, Raghda and Raghebi, Zohreh and Banaei-Kashani, Farnoush and Halbower, Ann C and Vu, Tam},
  booktitle={Proceedings of the 14th ACM Conference on Embedded Network Sensor Systems},
  pages={230--244},
  year={2016}
}

@article{fanger1970thermal,
  title={Thermal comfort. Analysis and applications in environmental engineering.},
  author={Fanger, Poul O},
  journal={Thermal comfort. Analysis and applications in environmental engineering.},
  year={1970},
  publisher={Copenhagen: Danish Technical Press.}
}

@book{mckeachie2006teaching,
  title={McKeachie’s teaching tips: Strategies, research, and theory for college and university teachers (12th ed.)},
  author={W. J. McKeachie and M. Svinicki},
  year={2006},
  publisher={Houghton-Mifflin}
}

@inproceedings{stirapongsasuti2019decision,
  title={Decision making support for privacy data upload in smart home},
  author={Stirapongsasuti, Sopicha and Sasaki, Wataru and Yasumoto, Keiichi},
  booktitle={Adjunct Proceedings of the 2019 ACM International Joint Conference on Pervasive and Ubiquitous Computing and Proceedings of the 2019 ACM International Symposium on Wearable Computers},
  pages={214--217},
  year={2019}
}

@inproceedings{papst2022share,
  title={To Share or Not to Share: On Location Privacy in IoT Sensor Data},
  author={Papst, Franz and Stricker, Naomi and Entezari, Rahim and Saukh, Olga},
  booktitle={2022 IEEE/ACM Seventh International Conference on Internet-of-Things Design and Implementation (IoTDI)},
  pages={128--140},
  year={2022},
  organization={IEEE}
}

@inproceedings{bovornkeeratiroj2020repel,
  title={Repel: A utility-preserving privacy system for iot-based energy meters},
  author={Bovornkeeratiroj, Phuthipong and Iyengar, Srinivasan and Lee, Stephen and Irwin, David and Shenoy, Prashant},
  booktitle={2020 IEEE/ACM Fifth International Conference on Internet-of-Things Design and Implementation (IoTDI)},
  pages={79--91},
  year={2020},
  organization={IEEE}
}

@ArtifactSoftware{R,
    title = {R: A Language and Environment for Statistical Computing},
    author = {{R Core Team}},
    organization = {R Foundation for Statistical Computing},
    address = {Vienna, Austria},
    year = {2019},
    url = {https://www.R-project.org/},
}

@inproceedings{elmalaki2019spycon,
  title={Spycon: Adaptation based spyware in human-in-the-loop iot},
  author={Elmalaki, Salma and Ho, Bo-Jhang and Alzantot, Moustafa and Shoukry, Yasser and Srivastava, Mani},
  booktitle={2019 IEEE Security and Privacy Workshops (SPW)},
  pages={163--168},
  year={2019},
  organization={IEEE}
}
